\documentclass[10pt,journal,compsoc]{IEEEtran}
\usepackage{booktabs}
\usepackage{algorithmic}
\usepackage[linesnumbered,ruled]{algorithm2e}
\usepackage{amsmath,amsfonts}
\usepackage{multirow}
\usepackage{makecell}
\usepackage{graphicx}
\usepackage{hyperref}
\usepackage{cite}
\hypersetup{hypertex=true,
            colorlinks=true,
            linkcolor=blue,
            anchorcolor=blue,
            citecolor=blue}
\usepackage[T1]{fontenc}

\ifCLASSINFOpdf
\else
\fi
\begin{document}
%
\title{Coordinated Sparse Recovery of Label Noise}
%
%
%
%

\author{Yukun~Yang,~\IEEEmembership{Student~Member,~IEEE},~Naihao~Wang,~Haixin~Yang,
        Ruirui~Li,~\IEEEmembership{Member,~IEEE}
\IEEEcompsocitemizethanks{\IEEEcompsocthanksitem Yukun Yang, Naihao Wang, Ruirui Li are with the Department of Information Science and Technology, Beijing University of Chemical Technology, Beijing. E-mail: yyk531177@163.com, 2022210508@mail.buct.edu.cn, ilydouble@gmail.com.
\IEEEcompsocthanksitem Haixin Yang is with the Department of Mathematical Sciences, Peking University. E-mail: yanghaixin527@pku.edu.cn.}
\thanks{Manuscript received April 19, 2005; revised August 26, 2015.}}

%
%

\markboth{Journal of \LaTeX\ Class Files,~Vol.~14, No.~8, August~2015}%
{Shell \MakeLowercase{\textit{et al.}}: Bare Demo of IEEEtran.cls for Computer Society Journals}
%



\IEEEtitleabstractindextext{%
\begin{abstract}
Label noise is a common issue in real-world datasets that inevitably impacts the generalization of models. This study focuses on robust classification tasks where the label noise is instance-dependent. Estimating the transition matrix accurately in this task is challenging, and methods based on sample selection often exhibit confirmation bias to varying degrees. Sparse over-parameterized training (SOP) has been theoretically effective in estimating and recovering label noise, offering a novel solution for noise-label learning. However, this study empirically observes and verifies a technical flaw of SOP: the lack of coordination between model predictions and noise recovery leads to increased generalization error. To address this, we propose a method called Coordinated Sparse Recovery (CSR). CSR introduces a collaboration matrix and confidence weights to coordinate model predictions and noise recovery, reducing error leakage. Based on CSR, this study designs a joint sample selection strategy and constructs a comprehensive and powerful learning framework called CSR+. CSR+ significantly reduces confirmation bias, especially for datasets with more classes and a high proportion of instance-specific noise. Experimental results on simulated and real-world noisy datasets demonstrate that both CSR and CSR+ achieve outstanding performance compared to methods at the same level.
\end{abstract}

}

\maketitle

\IEEEdisplaynontitleabstractindextext

%
\IEEEpeerreviewmaketitle

\IEEEraisesectionheading{\section{Introduction}\label{sec:introduction}}

%
%
%
%
\IEEEPARstart{D}{eep} learning has achieved remarkable success in various fields such as computer vision and medical image processing, primarily relying on high-quality labeled training data. Unfortunately, manually annotating data is costly and time-consuming, inevitably resulting in the introduction of noisy labels due to human bias and lack of expert knowledge. Alternatively, using web scraping or crowdsourcing to acquire labeled data often leads to a high proportion of label noise. Training neural networks directly on noisy data yields deteriorated performance, since over-parameterized deep neural networks have enough capacity to fit even completely random labels\cite{zhang2021understanding},\cite{arpit2017closer},\cite{jiang2018mentornet}. Consequently, the research area known as learning with noisy labels (LNL) \cite{song2022learning}, which seeks to address the adverse impact of label noise, has attracted considerable attention in recent years \cite{natarajan2013learning},\cite{manwani2013noise},\cite{van2015learning},\cite{ghosh2017robust},\cite{zhang2018generalized},\cite{wang2019symmetric},\cite{ma2020normalized},\cite{wang2022scalable}.

Some regularization methods \cite{Bishop1995tikhonov}, \cite{gu2008generative}, \cite{hoffman2019robust} aim to prevent overfitting to noise by imposing additional constraints, but this may introduce permanent bias. Noise transition matrix can be employed to represent the transfer relationship from clean labels to noisy labels. Although methods based on noise transition matrix have theoretical statistical consistency, accurately estimating them is a highly challenging task \cite{xia2020part}, \cite{yang2021estimating}, especially for instance-dependent label noise. Recent research has explored the use of memory effects through early learning methods\cite{jiang2018mentornet},\cite{zhang2018generalized},\cite{arazo2019unsupervised},\cite{cheng2020learning},\cite{englesson2021generalized},\cite{han2018co},\cite{li2020dividemix},\cite{karim2022unicon},\cite{li2021learning},\cite{liu2020early},\cite{tanaka2018joint},\cite{yu2019does},\cite{zhang2021learning} in LNL tasks. The most widely used approach is sample selection, which typically utilizes the confidence of the early-stage deep neural networks (DNNs) to select reliable samples for network update. However, these methods not only require prior knowledge of the noise-label ratio but also have the potential to select incorrect samples, leading to varying degrees of confirmation bias. Ways to make better use of early learning to reduce confirmation bias need further exploration.

In recent years, there has been a significant focus on the investigation of learning behavior in over-parameterized networks. Research \cite{zhu2022robustness} has discovered that wider over-parameterized networks demonstrate increased robustness, whereas the robustness of deeper over-parameterized networks relies on specific initialization methods and learning rates. In order to enhance early learning and unleash its potential, recent studies have introduced a robust training approach known as sparse over-parameterization (SOP). This approach leverages the inherent generalization ability of over-parameterization and integrates implicit regularization techniques to model and recover noisy labels that may differ across instances. Theoretical analyses demonstrate that SOP is effective in handling instance-dependent noise without introducing additional bias. However, SOP is beset by technical flaws and suffers from a non-coordinated learning issue between model prediction and label noise recovery. This issue results in error leakage, leading to overfitting and the memorization of noise, ultimately compromising the network's generalization capability.


This study analyzes this phenomenon and empirically concludes the following findings: 1) Sparse recovery in over-parameterized training exhibits a general lack of coordination between the learning of model parameters and that of noise parameters. 2) This lack of coordination contributes to error leakage and increased model prediction errors. Building upon these findings, we present a novel approach for coordinating sparse noise recovery through a collaboration matrix. It also suggests tuning the balance between the updates of model parameters and those of noise parameters by leveraging confident weights, consequently enhancing the reliability of model predictions. As a result, we propose the Coordinated Sparse Recovery (CSR) method. Comparing CSR with SOP, CSR effectively mitigates the issue of error memorization resulting from insufficient coordination and enhances generalization performance. Moreover, a joint sample selection strategy is introduced, based on the loss distribution and noise probability distribution derived during the CSR learning process. Constructing a robust learning framework using this strategy significantly mitigates confirmation bias and improves test accuracy. Our approach achieved superior performance on datasets with diverse proportions of instance-dependent noise, as well as on challenging real-world noisy datasets.

In summary, this paper presents several contributions, including but not limited to:

\hangafter 1
\hangindent 1em
\noindent
$\bullet$ Introducing the \textbf{collaboration matrix} to address the issue of non-coordinated learning between model prediction and label noise recovery, thus reducing the leakage of error gradients.

\hangafter 1
\hangindent 1em
\noindent
$\bullet$  Proposing the use of \textbf{confidence weight} to tune the balance between model parameter updates and sparse noise parameter updates, thereby improving the reliability of model predictions.

\hangafter 1
\hangindent 1em
\noindent
$\bullet$ Developing a novel \textbf{joint sample selection strategy} based on the loss distribution and noise probability distribution obtained from the Coordinated Sparse Recovery  (CSR), capable of partitioning samples into clean, hard, and noisy subsets.

\hangafter 1
\hangindent 1em
\noindent
$\bullet$ Conducting comprehensive experiments on synthetic instance-dependent noise datasets, as well as real-world noise datasets, to demonstrate the effectiveness of the proposed method.

\section{RELATED WORK}
\subsection{Learning From Noisy Labels}
We focus on discussing methods that do not require additional clean labeled datasets, nor consider methods for noisy label learning on open sets\cite{wei2021open}. 

The method designed for learning with noisy labels can be roughly summarized into three categories: robust loss functions, loss correction methods, and sample selection methods. 

Robust loss functions are designed to maintain robustness in the presence of label noise by using a cost function. This category of methods includes MAE loss \cite{ghosh2017robust}, improved reweighted MAE loss \cite{wang2019imae}, general cross-entropy loss \cite{zhang2018generalized}, symmetric cross-entropy loss \cite{wang2019symmetric}, and loss functions designed based on information theory like $L_{DIM}$ \cite{xu2019l_dmi}. Loss correction methods explicitly adjust the loss function to consider the noise distribution represented by a transition matrix of error probabilities \cite{goldberger2016training},\cite{xia2019anchor},\cite{patrini2017making},\cite{tanno2019learning}. The transition matrix has statistical consistency in theory, but it is difficult to accurately estimate them in the case of instance-dependent noise. The existing techniques of robust loss and loss correction have not fully explored the phenomenon of early learning\cite{arpit2017closer}, that is, deep convolutional neural networks tend to fit clean annotations before memorizing erroneous labels. Early learning can be explored through sample selection, where the outputs of these models in the early learning stage are used to predict whether the samples are incorrectly or correctly labeled, assuming that samples with erroneous labels tend to have higher loss values \cite{jiang2018mentornet,malach2017decoupling}.

\subsection{Sample Selection Methods}
Co-teaching \cite{han2018co} employs two parallel networks to select samples and update one network using the gradients of the samples with small-loss samples from the other network, in order to reduce error accumulation. However, this selection process may lead to the selection of overly simple and non-diverse samples, thus affecting the model's ability to generalize. Another approach involves correcting labels using models obtained during the early stage of learning. This is where Co-correcting \cite{liu2021co} comes into play, generating pseudo-labels for noisy samples by blending the probabilities of the one-hot labels with the model's prediction probabilities. DivideMix \cite{li2020dividemix} uses two Gaussian mixture models to estimate the predictive losses of two networks and selects samples based on this, combined with the MixMatch \cite{berthelot2019mixmatch} semi-supervised learning framework. To mitigate confirmation bias, many studies design optimized sample selection strategies. For instance, BARE \cite{patel2023adaptive} leverages an adaptive sample selection technique that statistics the samples within a batch. UNICON \cite{karim2022unicon} introduces a uniform selection mechanism based on JS divergence in the embedding space. DISC \cite{li2023disc} utilizes a dynamic and instance-specific threshold for sample selection.

In contrast to the explicit sample selection methods mentioned above that rely on specific strategies, the proposed CSR in this paper can be viewed as a loss correction method that dynamically explores and utilizes early learning characteristics. Through the collaboration matrix and confident weight, it achieves coordinated and reliable implicit sample selection and label correction.
\subsection{Sparse Over-Parameterization and its Variants}
The learning behavior of over-parameterized networks has garnered significant attention from researchers, particularly in light of the remarkable success of deep neural networks. Deep neural networks, being an archetypal case of over-parameterization, exhibit a phenomenon called double descent \cite{nakkiran2021deep}. This phenomenon is characterized by an initial decrease in test error, followed by an increase, and eventually a gradual decrease again. Some studies \cite{wen2022benign,nguyen2022memorization} suggest that this double descent phenomenon is associated with label noise, which aligns with the observations in the early learning stage.

Sparse over-parameterization \cite{liu2022sop} can be viewed as an implicit regularization technique. Unlike explicit regularization methods that introduce persistent bias, sparse over-parameterization exploits the inclination of over-parameterized networks towards low-rank and sparse solutions to separate and recover corrupted labels. Initially applied in the domain of image restoration \cite{you2020robust}, this approach can restore noisy images without any prior knowledge about the noise. In the realm of image classification, if we assume that the noise is sparse and dissimilar to the network trained on clean labels, we can employ the same idea to recover the labels. Subsequent research \cite{wani2023combining} has utilized this regularization term to train robustly and calculated the distance between noisy samples and class centers in the embedding space to generate pseudo labels. However, until now, there has been no investigation into the generalization degradation resulting from the lack of coordination of over-parameterized training with noisy labels.

\section{Preliminary}
\label{sec3}
Sparse Over-Parameterization (SOP) is a principled approach that aims to identify and recover sparse corrupted labels for each sample by adding a noise parameter vector $s_{i}$. It corrects the loss by adding the noise parameter vector rather than multiplying it with the noise transition matrix. Consequently, robust learning under noisy labels can be formulated as the following empirical risk minimization problem:
\begin{equation}
\label{e1}
\min\limits_{\theta,\{u_i,v_i\}_{i=1}^N}\frac{1}{N}\sum_{i=1}^N l(f(x_i;\theta)+s_i,y_i).
\end{equation}

Let $f$ denote the estimated probability of parameter $\theta$ given the input $x_i$, $s_i$ represent the noise vector, $y_i$ be the one-hot encoding of the label for the $i$-th sample, $N$ be the total number of samples, and $l$ denote the loss function. The aim of SOP is to utilize the previously learned robust model in the early stage to update $s_i$, and subsequently utilize the learned $s_i$ to mitigate overfitting of noise in the classification model. Given parameter $s_{i}$, denoting sparse noise, initialized with a negligible value near zero, and employing an optimal learning rate during the Stochastic Gradient Descent (SGD) for parameter updates, it becomes feasible to restore the noise parameter $s_i$ under the over-parameterization assumption with implicit regularization. The proof of this theory is detailed in reference \cite{liu2022sop}.

The transformation of $s_i$ can be expressed by the following formula, in order to decouple the instance-dependent label noise into non-negative vectors $u_{i}\odot u_{i}$ for identifying the noise and $v_{i}\odot v_{i}$ for recovering the noise:
\begin{equation}
\label{e2}
s_{i}=u_{i}\odot u_{i} \odot y_{i}-v_{i}\odot v_{i}\odot(1-y_{i}).
\end{equation}

In this context, $\odot$ is the Hadamard product. The vector $u_{i}$ is proportional to the one-hot label $y_{i}$, while the vector $v_{i}$ is associated with $(1-y_{i})$.

SOP utilizes the Cross-Entropy (CE) loss function for updating the parameters $\theta$ and $u$, and employs the Mean Squared Error (MSE) loss function for updating the parameter $v$. Prior to calculating the CE loss, SOP applies normalization to the sum of $f(x_i;\theta)$ and $s_i$. On the other hand, when learning the parameter $v$ with the MSE loss, SOP applies the one-hot transformation to $f(x_i;\theta)$, which promotes a more cautious learning approach.

\section{Coordinated Sparse Recovery }

\subsection{Analysis of the Learning Pace}
\label{sec4subA}
As expressed in Eq. \ref{e2}, the transition of the predicted probability of noisy labels can be decoupled into two components: adding the label noise probability and subtracting the predictions of the true labels. As a result, the SOP approach necessitates simultaneous learning of the network parameters $\theta$, and noise vectors $u$ and $v$ within the same network. If the learning pace of $u$ or $v$ is not synchronized with that of $\theta$, it may introduce substantial bias into the transition process, causing the inclusion of inaccurate information in the gradient update of $\theta$ and resulting in erroneous gradient leakage. This leakage subsequently impedes the proper learning of $\theta$, leading to more pronounced error accumulation.

To analyze the learning paces of these three groups of parameters, we computed the distribution of their gradient updates throughout the training. The proportion of gradient updates indicated as $g_t$ at the $t$-th epoch, can be expressed as:
\begin{equation}
\label{e3}
g_t=\frac{|z_t|}{\sum^T_{t=1}|z_t|},t=1,2,\cdots T.
\end{equation}

\begin{figure*}
\centering
\includegraphics[width=7in]{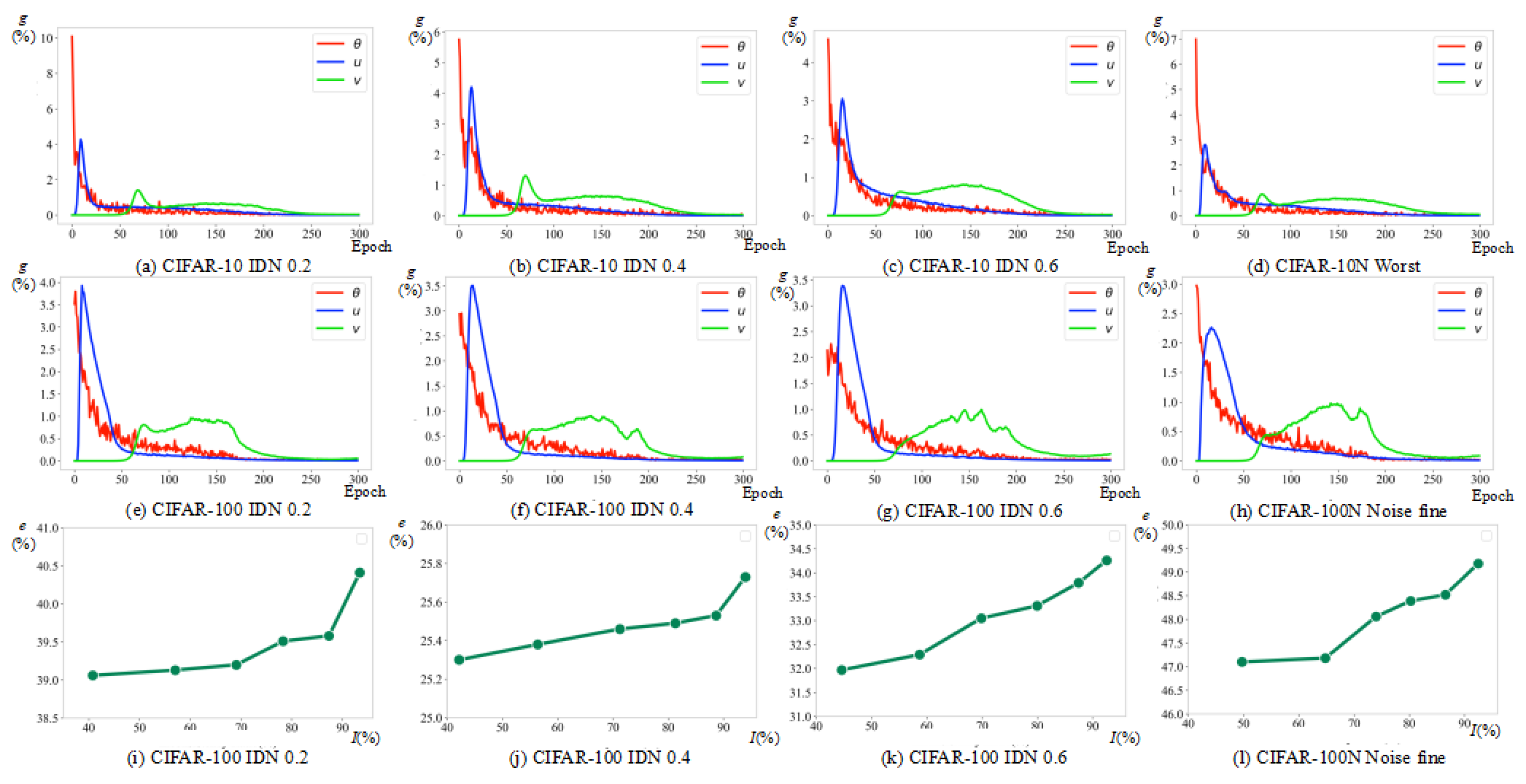}
\caption{ 
Empirical analysis of the incoordination phenomenon of $v$-lag learning and its impact on test error on CIFAR-10/100 Datasets with different label noises: variation trends of parameter gradient with respect to training time (a)$\sim$(h) and correlation between parameter learning incoordination and model generalization (i)$\sim$(l).
}
\label{fig2}
\end{figure*}

A higher value of $g_t$ implies a stronger engagement in the learning tasks during the $t$-th epoch. $|z_t|$ signifies the cumulative sum of absolute gradients for the parameter set. $T$ represents the total number of epochs. Figure \ref{fig2}a$\sim$h exhibits the distribution of gradient update for different parameter groups when trained on CIFAR-10/100 with varying ratios of instant-dependant noisy labels using the hyperparameters given in reference\cite{liu2022sop}.

It is apparent that, regardless of different datasets and label noise ratios, a consistent trend emerges: the learning of $\theta$ and $u$ show substantial overlap, while the learning of $v$ noticeably falls behind $\theta$ and $u$. This suggests that during the learning processes of $\theta$ and $u$, there exists a significant duration where $v$ remains unchanged at a negligible level. Within this timeframe, if $f(x_{i},\theta)$ exhibits considerable confidence for a specific term in the vector $(1-y_{i})$,  it will generate an erroneous gradient updating in the opposite direction. Consequently, the model's confidence in the correct category will then decrease, causing fluctuations back and forth.

 The one-hot operation plays a key role in causing the delayed learning of $v$. In situations where mislabeled samples are present and the model is uncertain about selecting the correct class, applying the one-hot transformation will result in all predictions for classes other than the labeled class $y_i$ being zero. This leads to a zero gradient for updating $v$ at a specific time point. Only when the model becomes confident enough in selecting the correct class, it will generate gradients that can update $v$, along with gradient errors that update $\theta$. In fact, the one-hot operation is crucial for the SOP algorithm, and removing this transformation would result in a significant increase in the model's test error. The delay in learning $v$ can also be attributed to using the flatter Mean Squared Error (MSE) loss function instead of the Cross-Entropy (CE) loss update. Appendix B offers a detailed derivation and explanation as to why the adoption of the CE loss function is not feasible, despite the fact that utilizing MSE further worsens the challenge of coordinating the noise parameter $v$ and the model parameter $\theta$.

\subsection{Influence to Generalization }
\label{sec4subB}
This paper examines the extent of incoordination in learning paces by employing the Chi-Square\cite{chi-square} distance to quantify the disparity in gradient distributions. Given that $g^\theta$ and $g^v$ denote the gradient distributions of two distinct parameter groups, and $g^\theta_t$ and $g^v_t$ denote the proportion in the $t$-th epoch respectively, we define the ratio of incoordination as follows:
\begin{equation}
\label{e4}
I=\frac{\sum^{T}_{t=1}|g^\theta_t-g^v_t|}{\sum^{T}_{t=1}g^\theta_t+g^v_t}.
\end{equation}

When the distributions of the gradients are perfectly identical, the value of $I$ is $0$. Conversely, when there is no overlap at all between the distributions of the gradients, the value of $I$ is $1$.

This research conducts approximate simulation experiments to analyze the association between the incoordination ratio $I$ and the test error $e$. The experiments are carried out on the  CIFAR-10/100 datasets, with varying ratios of instance-dependent noise as well as real-world noise. The experiment setup involves training the network initially and storing the values and gradients of $v$ for each epoch. Subsequently, 30 epochs of $v$ are repeatedly shifted backward, where after each shift, the update of $v$ remains fixed, and the values of $v$ are utilized for training the model parameters $\theta$ and sparse parameters $u$. Then, the testing errors of the model are computed after each shift.

Figure \ref{fig2}i$\sim$l showcases partial experimental findings on the CIFAR-100 dataset, incorporating various ratios of instance-dependent noise and real-world noise. The $x$-axis denotes the level of incoordination $I$, while the $y$-axis represents the test error $e$. It can be observed that the change curve exhibits a monotonic increase, indicating that as the value of $I$ augments, the test error rises. Accordingly, the experiment affirms that the incoordination ratio is in fact linked to the test error of the model, thereby compromising its generalization capability.

\subsection{Collaboration Matrix}
\label{sec4subC}
In order to tackle the problem of learning incoordination during over-parameterized training, this paper introduces an innovative approach, namely the collaboration matrix, which serves to dynamically counteract error leakage. The label estimation process is modeled as the multiplication of predictions by the collaboration matrix, followed by the addition of a sparse noise vector. More specifically, we redefine the optimization problem for empirical risk minimization in the following manner:
\begin{equation}
\label{e5}
\min\limits_{\theta,\{u_i,v_i\}_{i=1}^N}\frac{1}{N}\sum_{i=1}^N l(f(x_i;\theta)\times M+s_i,y_i)
\end{equation},
where $M\in R^{K\times K}$ is the collaboration matrix. The matrix $M$ is learnable. It is initialized with the identity matrix and updated using stochastic gradient descent along with other parameters. The CE loss is employed to update the $M$ matrix, and the specific update formula is as follows:
\begin{equation}
\label{e6}
M\longleftarrow M-\tau_m \cdot\frac{\partial L_{CE}}{\partial M},i=1,\dots,N,
\end{equation}
where $\tau_{m}$ represents the learning rate of the collaboration matrix $M$. We employ loss calculation and gradient update methods analogous to SOP. The distinction lies in the inclusion of matrix $M$ in the calculation and update processes. Moreover, to mitigate issues of gradient explosion and gradient disappearance during training, we additionally apply matrix normalization. This is demonstrated in the formula provided below.

\begin{equation}
\label{e50}
\bar{M}=(M-min(M))/(\gamma-min(M)),
\end{equation}
in which $\bar{M}$ is the normalized matrix and $\gamma$ is a learnable scale parameter.

\begin{figure}[h]
\centering
\includegraphics[width=3.5in]{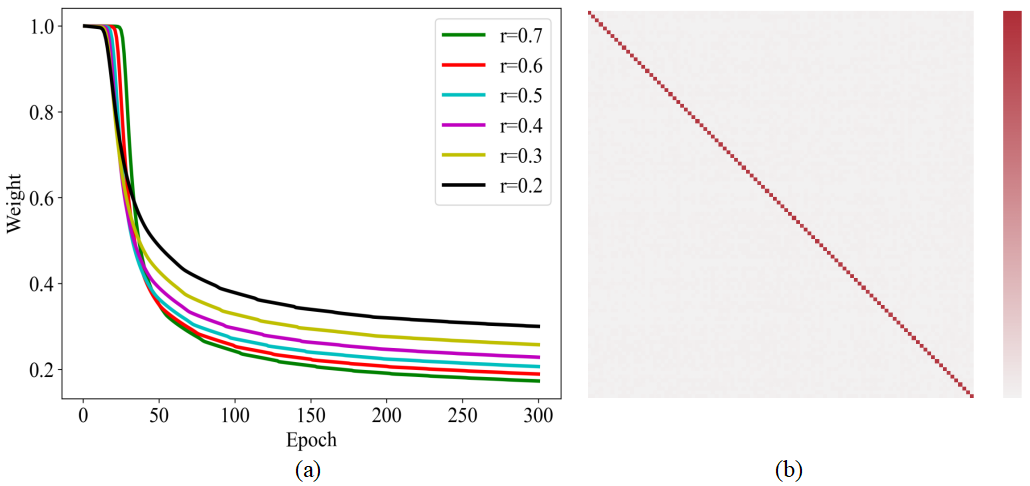}
\caption{Visualization and change analysis of M matrix values: (a) average value variations in M matrix over time; (b) visualization of M value at 50 epoch time: mainly concentrated on the diagonal}
\label{fig4}
\end{figure}

Within the training of this research, $M$ primarily focuses on updating the parameters located on the main diagonal of the matrix, whereas the other off-diagonal parameters tend to approach $0$. Figure 
 \ref{fig4}b provides a visual representation of $M$ after 50 epochs under 40\% IDN noise on CIFAR-100. From this figure, we can easily observe this phenomenon. Our speculation is that this behavior may stem from the distinct update stages of the decoupled orthogonal vectors $u$ and $v$, which are responsible for noise prediction and noise correction respectively in the context of SOP. This disparity also allows $M$ to play a collaborative coordinating role in $f$. Specifically, during the period of delayed learning for $v$, multiplying $f$ by $M$ effectively narrows the gap for mislabeled samples, thereby reducing the contamination of incorrect gradients.

We also analyze the fluctuation of $M$. Figure \ref{fig4}a illustrates the average value on the main diagonal of $M$ obtained through training on  CIFAR-100 with various ratios of instance-dependent noise. Notably, these curves demonstrate a consistent trend followed by a rapid decline starting from an initialization value of $1$. Particularly, during the phase of delayed learning for vector $v$, the mean value of $M$ experiences a sharp decrease. As $v$ begins to update, the mean value of $M$ undergoes a gradual and slow decline until it reaches a stable state.

\begin{figure*}
\centering
\includegraphics[width=5in]{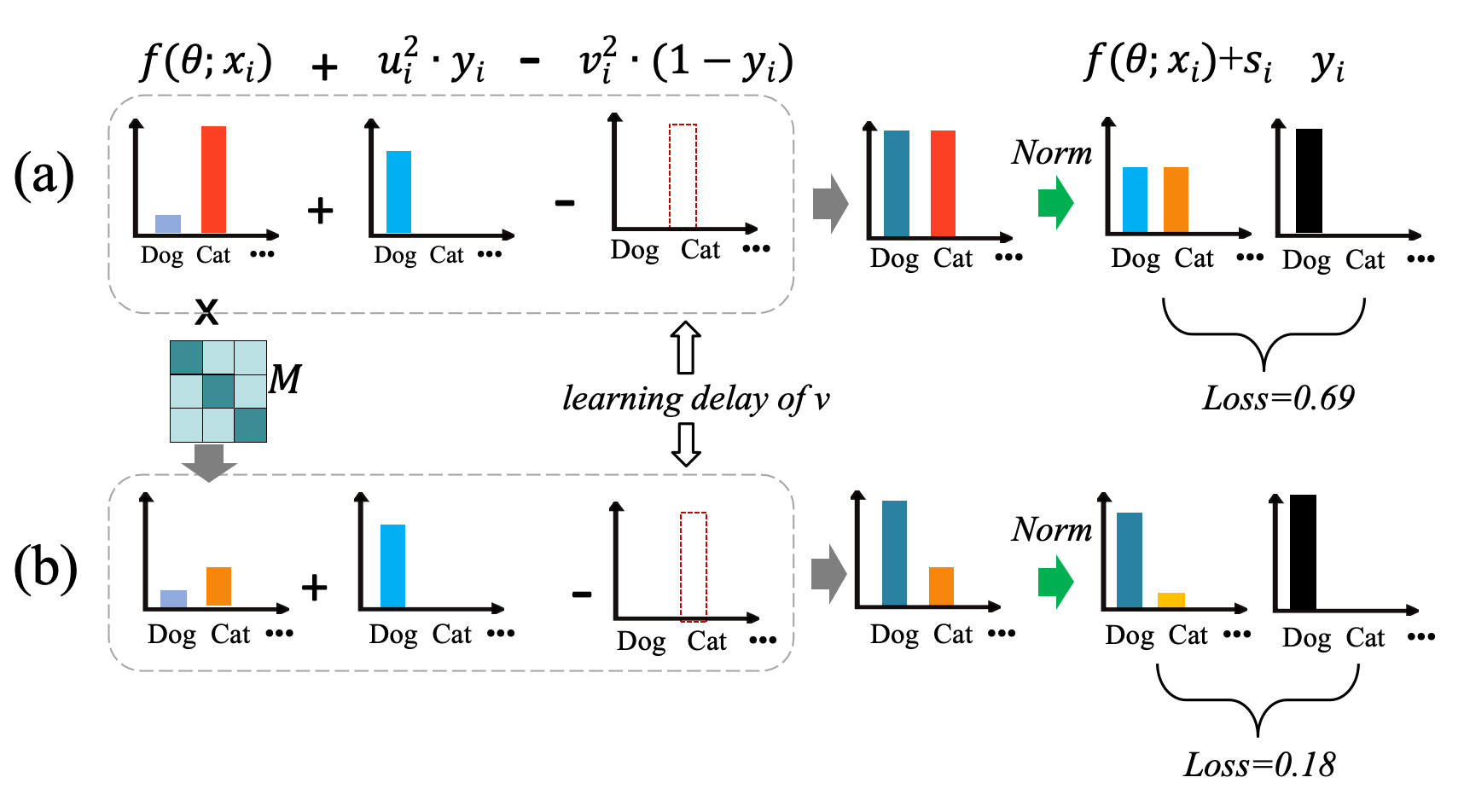}
\caption{Comparison on an incorrectly annotated sample (b) with and (a) without the use of a collaboration matrix. Without the collaboration matrix, there would be an error loss of 0.69, whereas, after the utilization of the collaboration matrix, the error loss decreases significantly to 0.18.}
\label{fig5}
\end{figure*}

We adopt a unified $M$ for all samples in our learning process. Each element on the diagonal of $M$ is approximated to represent the adjusted weight for its corresponding category. Figure \ref{fig5} illustrates the learning process for mislabeled samples, highlighting the difference between using $M$ and not using it. When $M$ is not applied, the predicted probability $f$ for the correct category cannot be effectively corrected by $v$, resulting in a significant loss that should ideally be avoided. However, by multiplying $M$ with the predicted probability, this erroneous loss is greatly reduced. This helps in mitigating gradient updates in the wrong direction and slowing down the learning of $\theta$. It is worth noting that $M$ does not significantly affect the loss computation for correctly labeled samples. In cases where $s_{i}$ tends close to 0, only the product of $M$ and $f$ needs to be computed, which undergoes minimal change after normalization.

\subsection{Confidence Weight }
\label{sec4subD}
To further improve the robustness for $\theta$, we introduce a technique called confidence weighting that adjusts the learning proportion for different groups of parameters. Represented as $\omega$, the weight is a floating-point value ranging from 0 to 1. We multiply $\omega$ with the gradient of the model parameters and $(1-\omega)$ with the gradient of the noise parameters. Our assumption is that when the predicted probability closely matches the given label, the confidence weight tends to be $1.0$. This indicates that the sample is likely to be accurately labeled and justifies updating the model parameters. Conversely, when the confidence weight approaches 0, it suggests that the sample is likely mislabeled and requires an update to the noise parameters.

To ensure more accurate predicted probabilities, we employ a momentum-based exponential moving average (EMA) on the results of $\Delta T$ time-series predictions. Suppose $t$ is the current epoch,  $q'_t$ denotes the model prediction at $t$, and $q_t$ is the temporarily ensembled prediction and adopts the following formula for the recursive operation.
\begin{equation}
\label{e7}
q_t=\beta \cdot q_{t-1}+(1-\beta_t)\cdot q'_t,
\end{equation}
\begin{equation}
\label{e8}
\beta_t=\frac{(\beta_{init}-1)\cdot t}{T}+1,
\end{equation}
where $\beta_t$ represents the momentum, which is dynamically computed to prioritize the predictions of the model at a nearer time and ensure that the smoothed outcome reflects this consideration more prominently.  $\beta_{init}$ is a hyperparameter, and $T$ is the overall number of training epochs.

Let $p_i=q_{\Delta T}$ denote the predicted probability after applying temporal EMA smoothing over $\Delta T$ windows. The confidence weight can be computed using the following formula.
\begin{equation}
\label{e9}
\begin{split}
\omega_i=1-Normalize_{[0,1]}(L_{CE}(p_i,y_i)), \\
\Omega=\{\omega_i; i\in (1,\dots,N)\}, \ \ \ \ \ \ \ \ \ \ 
\end{split}
\end{equation}
where $L_{CE}$ is the cross-entropy loss, $y_i$ is the given label, and $\Omega$ represents the set of all training samples.

\subsection{Overall Algorithm of CSR}
\label{sec4subE}

\begin{figure}[h]
\centering
\includegraphics[width=3in]{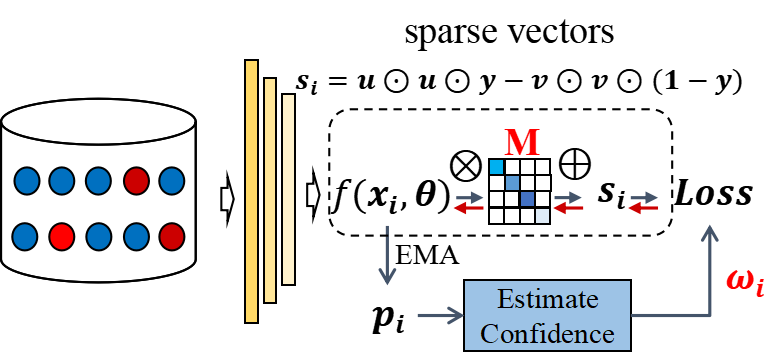}
\caption{The architecture of CSR.}
\label{fig6}
\end{figure}

\begin{figure}[h]
\centering
\includegraphics[width=3.4in]{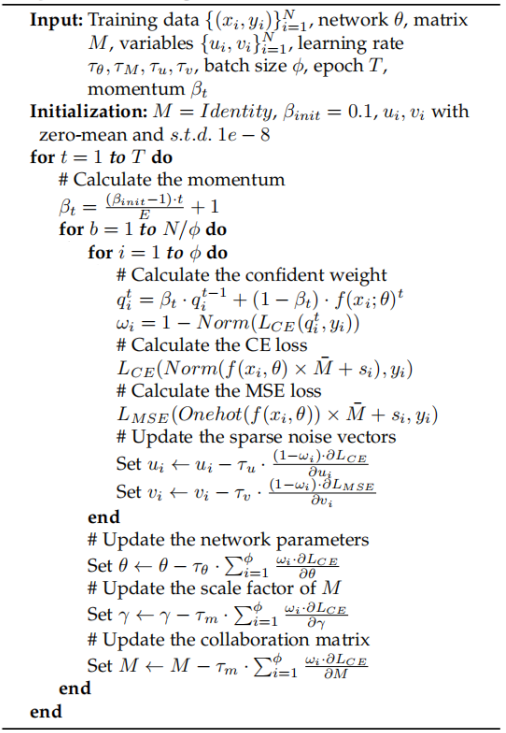}
\caption{Overall Algorithm of CSR.}
\label{alg}
\end{figure}

We put forward a resilient training approach for overparameterized networks, referred to as Coordinated Sparse Recovery (CSR), that revolves around the collaboration matrix and the confidence weighting. The configuration of CSR is depicted in Figure \ref{fig6} and the overall algorithm is shown in Figure \ref{alg}.

In each batch, we update the network parameters $\theta$, the scale factor $\gamma$, and the collaboration matrix $M$. In each epoch, we update $u_i$ and $v_i$ for each sample. The emphasis of the updates is determined by calculating the confidence weight $\omega_i$ of each sample.

\section{Extension of CSR}
CSR can serve as a standalone powerful learning method and also as a component in combination with semi-supervised techniques to devise a novel sample-selection-based learning framework.

\subsection{Small-$u$ Sample Selection}
\label{sec5subA}
In sample selection, it is commonly employed to utilize a Gaussian mixture model to estimate the distribution of sample losses, followed by picking out clean samples with small losses using a given threshold, denoted as $\sigma$. Let $GMM()$ represent the process of fitting a Gaussian mixture model, and $L_{CE}(f(x_i), y_i)$ the cross-entropy loss between the prediction $f$ and its corresponding label $y_i$ for a sample $x_i$. The set of selected clean samples can be computed using the following formula.
\begin{equation}
\label{e10}
\begin{split}
S_{loss}=\{x_i,y_i|GMM(L_{CE}(f(x_i),y_i))>\sigma\},
\end{split}
\end{equation}

The precision of sample selection by this strategy may not be optimal, as difficult-to-learn samples also have big losses. On the other hand, during CSR learning, we utilize $u$ to model the probability of label noise, hence the distribution of $u$ can serve as a criterion for sample selection. To address this, we propose the small-$u$ strategy. Specifically, we use $GMM$ to fit the distribution of $u$, and then apply a threshold cutoff on the results to filter out clean samples, which is defined as follows:
\begin{equation}
\label{e11}
\begin{split}
S_{u}=\{x_i,y_i|GMM(u_i)>\sigma\},
\end{split}
\end{equation}

\begin{figure}[h]
\centering
\includegraphics[width=3.4in]{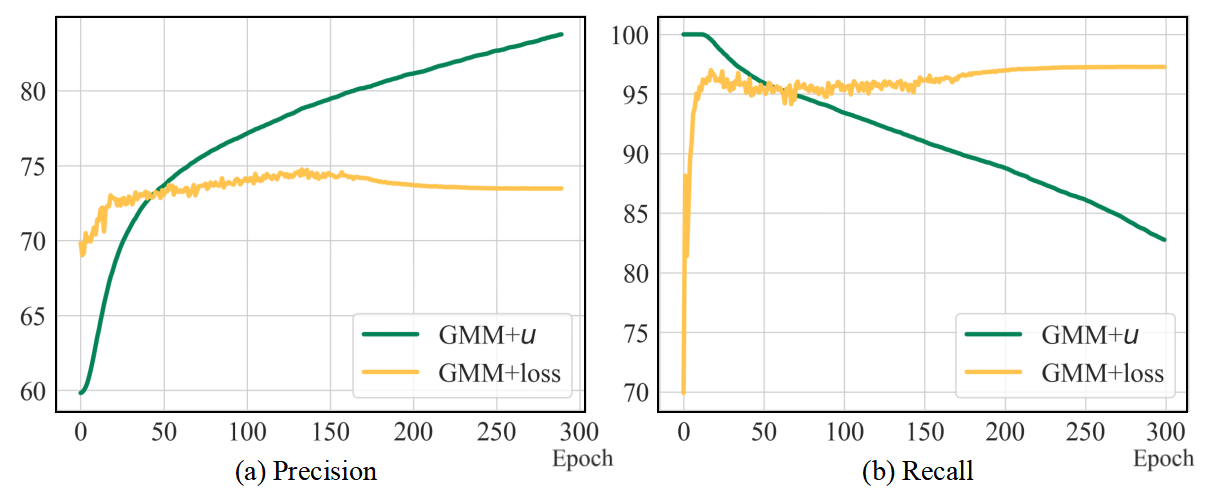}
\caption{Comparison of precision and recall for small-loss sample selection and small-$u$ sample selection.}
\label{fig8}
\end{figure}

Figure \ref{fig8} demonstrates the selection performance on the CIFAR-100N dataset with real-world label noise, employing the small-loss strategy and small-$u$ strategy respectively. As depicted in Fig. \ref{fig8}a, the precision of samples chosen via GMM, fitted to the $u$-distribution, consistently improves with longer training time, eventually reaching a precision of more than 90\%. Nevertheless, the recall of the small-$u$ selection exhibits a perpetual decline,  as shown in Fig. \ref{fig8}b, unlike the stable recall observed in the small-loss sample selection strategy.  

\subsection{Joint Sample Selection}
This paper presents a methodology for conducting joint sample selection. By utilizing set operations, the complete sample dataset is ultimately divided into three well-defined subsets, namely the clean set, hard set, and noisy set. This approach aims to simultaneously address both accuracy and diversity concerns in sample selection. Given that $S_{loss}$ is derived from Eq. \ref{e10} and $S_u$ is derived from Eq. \ref{e11}, the computation methods for the clean set $S_{clean}$, the hard set $S_{hard}$, and the noisy set $S_{noisy}$ can be defined as follows in equations \ref{e12} to \ref{e14}.
\begin{equation}
\label{e12}
S_{clean}=\{S_{loss}\cap S_u\},
\end{equation}
\begin{equation}
\label{e13}
S_{hard}=\{S_{loss}\cup S_u-S_{loss}\cap S_u\},
\end{equation}
\begin{equation}
\label{e14}
S_{noisy}=\{\Omega-S_{loss}\cup S_u\}.
\end{equation}

We employ different processing strategies for subsets of clean, hard, and noisy samples. Given the high reliability of sample labels in $S_{clean}$, we utilize supervised learning with both strong and weak augmentation, as well as Mixup augmentation. This approach aims to enhance the model's performance by leveraging trustworthy labels to their fullest extent. On the other hand, the samples in $S_{hard}$ exhibit better diversity but less reliability compared to $S_{clean}$. To address this, we only apply weak and Mixup augmentation with supervised learning. This strategy aims to achieve a more robust decision boundary by locally smoothing. As for the samples in $S_{noisy}$, their labels are considered unreliable, and thus, we treat them as unlabeled samples. To handle this, we generate pseudo-labels for the noisy samples using a dynamic threshold. We then include the samples with pseudo-labels into the clean set for the subsequent round of training. This process helps us progressively improve the model's performance on the noisy samples by gradually incorporating them into the training process.

\subsection{Framework of CSR+}
\label{sec5subc}
\begin{figure}[h]
\centering
\includegraphics[width=3.4in]{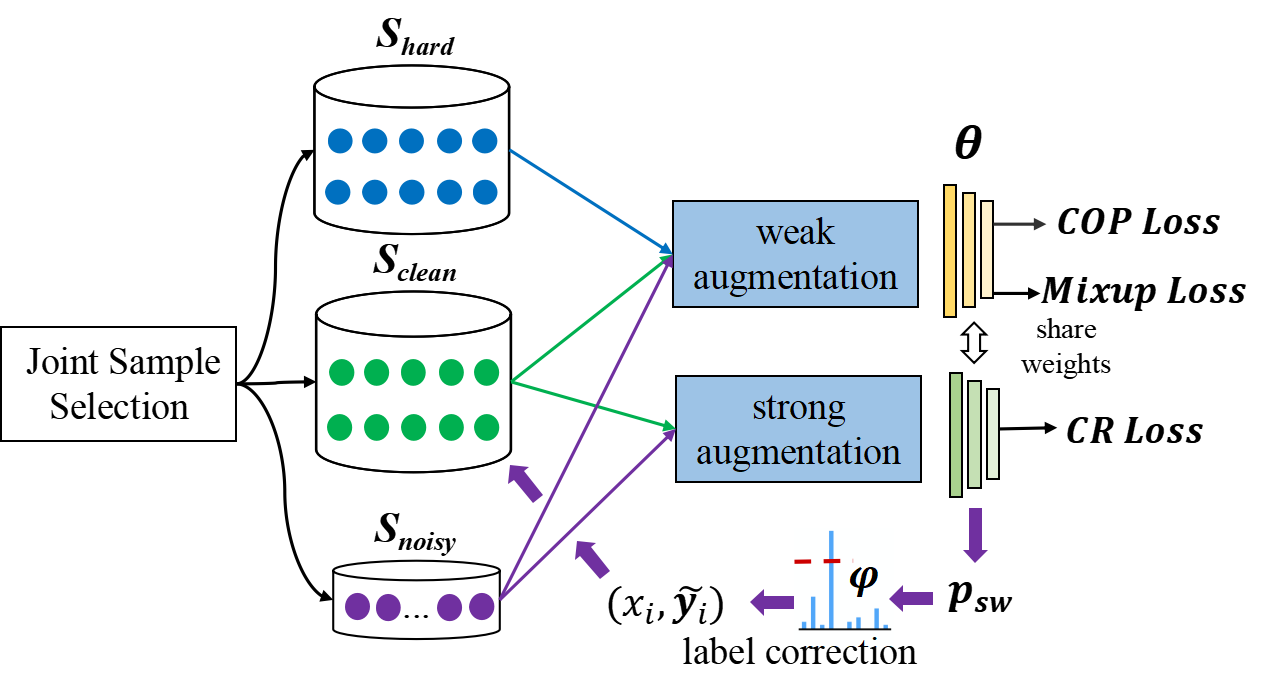}
\caption{CSR+ architecture.}
\label{fig10}
\end{figure}

The CSR+ framework is designed by integrating multiple techniques including joint sample selection, consistent regularization with strong and weak augmentation, Mixup augmentation, and label correction techniques. Figure \ref{fig10} depicts the architecture of CSR+, featuring a multi-task learning network with three losses. In order to provide a comprehensive understanding, this paper will briefly outline the technical aspects of consistent regularization, Mixup, and label correction.\\
\subsubsection{Consistency Regularization}

The consistency regularization used in the CSR+ framework is identical to the consistency regularization employed in the previous work {\cite{li2020dividemix},\cite{wang2022promix}}. It involves generating a new sample by applying random augmentations to the original sample, and then calculating the CE loss with its corresponding label. Let $x_i^s$ represent the augmented sample of $x_i$, and let $y_i$ denote its corresponding label. The consistency regularization can be calculated using the following formula: 

\begin{equation}
\label{e15}
L_{cr}=L_{CE}(f(x_i^s,\theta),y_i).
\end{equation}
\subsubsection{Mixup}

The CSR+ framework employs Mixup augmentation technique to generate new training instances using weakly augmented samples. Let $x_i^w$ and $x_j^w$ be two weakly augmented samples, and $y_i$ and $y_j$ be their corresponding labels. The newly generated training instance $\{x_i^m, y_i^m\}$ can be expressed as follows:
\begin{equation}
\begin{split}
\label{e16}
x_i^m=\delta \cdot x_i^w+(1-\delta)\cdot x_j^w, \\
y_i^m=\delta \cdot y_i+(1-\delta)\cdot y_j, \ \  
\end{split}
\end{equation}
where $\delta\in [0,1]$, $\delta\sim Beta(\alpha,\alpha)$ is a hyperparameter representing the mixing ratio. Then, the Mixup loss is defined as follows:
\begin{equation}
\label{e17}
L_{Mix}=L_{CE}(f(x_i^m,\theta),y_i^m).
\end{equation}
Until now, the overall loss can be calculated as follows:
\begin{equation}
\label{e18}
L_{overall}=L_{CSR}+\alpha(L_{CR}+L_{Mix}),
\end{equation}
where $L_{CSR}$ is the loss function for the training of CSR and $\alpha$ is a weight factor hyperparameter set to increase linearly from 0 to 1.\\

\subsubsection{Noisy Label Correction}

The CSR+ framework employs the dynamic threshold sample correction method proposed by DISC\cite{li2023disc} to rectify the samples within the set of noisy samples. This correction method consists of three steps.

\textbf{Step1}: Employ a weighted averaging technique to combine the predicted probabilities derived from the strong and weak augmented predictions.
\begin{equation}
\label{e19}
p_{ws}=\epsilon\cdot p_w+(1-\epsilon)\cdot p_s.
\end{equation}
The variables mentioned in the provided formula include the following: $p_w$, which represents weakly augmented prediction; $p_s$, which represents strongly augmented prediction; $p_{ws}$, which represents the newly obtained prediction probability; and $\epsilon$, indicating the weight coefficient set to 0.5.

\textbf{Step 2}: Calculate the dynamic threshold value.
\begin{equation}
\label{e20}
\phi(t)=max(\phi_{ws}(t)+0.5, 0.99),
\end{equation}
where 0.5 is a small positive offset value, 0.99 is an upper threshold and $\phi_{ws}(t)$ is the historical smoothing of $p_{ws}$

\textbf{Step 3}: Employ the dynamic threshold to filter the obtained new predicted probabilities, replacing the original labels with the highest predicted probability that surpasses the threshold. This revised probability will then serve as the pseudo-label.
\begin{equation}
\label{e21}
\tilde{y_i}=argmax(p_{ws})|min(p_{ws})>\phi(t).
\end{equation}

\section{ EXPERIMENT AND ANALYSIS }

To evaluate the effectiveness of CSR and CSR+, we conducted extensive experiments on synthetic datasets containing instance-dependent label noise at different noise ratios sourced from CIFAR-10/100. Additionally, we evaluated these models on five real-world datasets with label noise. While our primary objective is to enhance the model's robustness to instance-dependent label noise, it is noteworthy that our approach also exhibits significant efficacy in handling simpler types of label noise.

In this chapter, we begin by comparing CSR with other principle approaches (without using any additional training techniques) to evaluate the generalizability of CSR. Then, we discuss the contributions of the two core modules of CSR and the learning rate settings of the collaboration matrix in the training process. Following that, we compare CSR+ with other state-of-the-art methods to provide a comprehensive performance assessment. Subsequently, we conduct ablation studies to investigate the contributions of joint sample selection and other training techniques employed in CSR+. Lastly, we analyze the complexities of CSR and CSR+, including time complexity and space complexity.

\subsection{Datasets}
The experiments make use of the following datasets:

\hangafter 1
\hangindent 1em
\noindent
$\bullet$\ \ CIFAR-10/100 \cite{krizhevsky2009cifar100} are datasets comprising of 32 × 32 × 3 color images. Both of them consist of a total of 50,000 training images and 10,000 test images.

\hangafter 1
\hangindent 1em
\noindent
$\bullet$ CIFAR-10N/100N datasets \cite{wei2021cifar100n} are real-world noisy datasets that have been manually annotated. Each image in CIFAR-10N has three annotated labels. The label denoted as Random $i$ represents the $i$-th submitted label for each image. The label denoted as Aggregate is obtained through aggregating the three noisy labels by majority voting. The dataset with the highest noise rate is referred to as Worst. We have selected the Worst label from CIFAR-10N, which has a noise rate of 40.21\%. On the other hand, each image in CIFAR-100N is associated with a coarse label and a fine label, both of which are provided by human annotators. We have selected the noisy fine labels from CIFAR-100N, which exhibit an overall noise level of 40.20\%.

\hangafter 1
\hangindent 1em
\noindent
$\bullet$ Clothing1M\cite{xiao2015clothing1m} is a real-world dataset that comprises of 1 million training images extracted from online shopping sites. It consists of 14 categories. The estimated noise level in this dataset is 38.5\%.

\hangafter 1
\hangindent 1em
\noindent
$\bullet$ WebVision \cite{li2017webvision} and ILSVRC12 are datasets derived from the ImageNet Large Scale Visual Recognition Challenge. The ImageNet dataset contains 1,000 classes and 2.4 million images. In our experiments, we follow the Mini setting and use the data from the initial 50 classes for training. The estimated noise level in these datasets is 20\%. To evaluate our model, we utilize the validation sets of WebVision and ILSVRC12, which consist of 50,000 and 2,500 images respectively.

\subsection{Synthetic Noise}

To synthesize instance-dependent noise, we employ the method mentioned earlier \cite{xia2020part}, which depends on part-specific label noise to generate noise that is specific to each instance. Our experiments were carried out using three different noise rates, namely 20\%, 40\%, and 60\%.

\subsection{Experimental setup}
Experiments were conducted using a GeForce RTX3090 GPU in the PyTorch v1.10 environment. The SGD optimizer was utilized, and the learning rate settings were following those of SOP. The experimental setups for each dataset were as follows:

For the CIFAR-10/100 and CIFAR-10/100N datasets, the PreActResNet-18 network was employed as the backbone. Learning rate for this network was set to 0.02 for CSR and 0.03 for CSR+, with a weight decay of 0.0005 and a batch size of 128. The network was trained for 300 epochs. For the CIFAR-10 and CIFAR-10N datasets, the learning rate for the matrix was set to 0.000001, and a warm-up period of 10 epochs was applied. For the CIFAR-100 and CIFAR-100N datasets, The learning rate for the matrix was set to 0.001, and a warm-up period of 20 epochs was applied.

In the case of the Clothing1M dataset, a pre-trained ResNet50 served as the backbone network. The learning rate for this network was set to 0.002, with a weight decay of 0.001 and a batch size of 32. The learning rate for the matrix was set to 0.001, with a warm-up period of 1 epoch. The overall training process lasted for 10 epochs.

For the WebVision and ILSVRC12 datasets, the Inception ResNetv2 was used as the backbone network. The learning rate for this network was set to 0.01, with a weight decay of 0.01 and a batch size of 32. The learning rate for the matrix was set to 0.001, with a warm-up period of 5 epochs. The overall training process lasted for 80 epochs.
\subsection{Comparison of CSR With Other Principle Methods}

To begin with, we conduct a comparative analysis of CSR with several other principle methods, namely CE \cite{de2005ce} loss, Forward method \cite{patrini2017making} that integrates noise estimation and loss correction, Co-teaching \cite{han2018co} which adopts the dual network sample selection technique,  GCE \cite{zhang2018generalized} loss which combines CE loss and MAE loss\cite{willmott2005mae}, ELR \cite{liu2020early} with the early learning regularizer, and SOP \cite{liu2022sop} using sparse over-parameterized training.

\begin{table}[h]
\caption{Comparison of CSR With Other Principle Methods}
\label{table1}
\begin{center}
\setlength\tabcolsep{3.5pt}
\begin{tabular}{ccccccccc}
\toprule
\upshape{Dataset}& \multicolumn{3}{c}{CIFAR-10} & \makecell{10N}& \multicolumn{3}{c}{CIFAR-100} & \makecell{100N}\\
\midrule
 Noise type & \multicolumn{3}{c}{Instance dependent} & \makecell{Worst}&\multicolumn{3}{c}{Instance dependent} & \makecell{Noise \\ fine}\\
\midrule
Noise rate & 0.2 & 0.4 & 0.6 &\makecell{about \\ 0.4}& 0.2 & 0.4 & 0.6 & \makecell{about \\ 0.4}  \\
\midrule
CE &  83.93 & 67.64& 43.83 &77.69& 57.35 & 43.17 & 24.42 & 55.50 \\
Forward & 87.22& 79.37& 66.56 & 79.79& 58.19 & 42.80 & 27.91 & 57.01\\
Co-teaching & 88.87& 73.00& 62.51&83.83& 43.30 & 23.21 &	12.58 & 60.37\\
GCE &  89.80 & 78.95 & 60.76& 80.66& 58.01 & 45.69 & 35.08 & 56.73\\
ELR & 92.35	& 88.74&  67.97&83.58 & 72.66 & 62.04	& 44.38	& 59.81\\
SOP & 93.91&  91.95& 72.12&86.76& 74.09	& 66.52	& 50.20	& 61.09\\
CSR & \textbf{94.08}&\textbf{92.56}& \textbf{74.69}&\textbf{87.37}& \textbf{75.93}	& \textbf{71.75}	& \textbf{58.98}	& \textbf{64.62}\\
\bottomrule
\end{tabular}
\end{center}
\end{table}

\begin{figure}[h]
\centering
\includegraphics[width=3.6in]{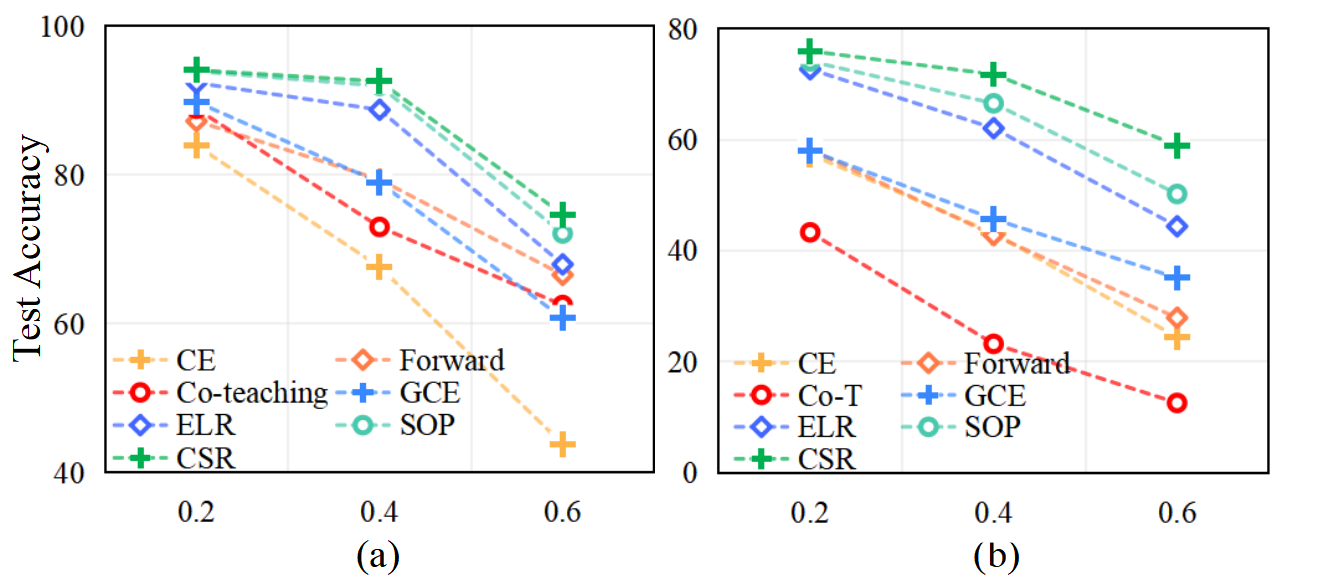}
\caption{Visualization of comparative results of principle methods on a) CIFAR-10 and b) CIFAR-100 with varying IDN ratios.}
\label{fig11}
\end{figure}

Table \ref{table1} presents a comprehensive analysis of the experimental results obtained by comparing CSR with the other approaches. To facilitate analysis and enhance understanding, these results have been graphically represented in a curve graph, as shown in Figure \ref{fig11}.  The CSR method demonstrates superior performance compared to other methods. As the noise ratio increases, the test accuracy of the comparative methods declines significantly. For instance, when tested on the CIFAR-10 IDN dataset, the classical Co-teaching method experienced a drop in test accuracy from 88.87 to 62.51. Similarly, the SOP method showed a decrease in test accuracy from 74.09 to 50.20 on CIFAR-100 IDN dataset. In contrast, the CSR method exhibited the smallest decline in test accuracy on both datasets, indicating its robustness. Further analysis consistently ranks the SOP method as the second-best in all metrics, closely following the CSR method. However, as the noise ratio increases, the performance gap between CSR and SOP widens. Moreover, the CSR method demonstrates more notable performance improvement on the CIFAR-100 dataset compared to CIFAR-10. When tested on CIFAR-10 with a 60\% IDN, CSR outperforms SOP with a 2.57 percentage point increase. Conversely, when tested on CIFAR-100 with a 60\% IDN, this improvement escalates to 8.78. These empirical findings undeniably establish the advantages of the CSR method, especially in more challenging scenarios.\\

\begin{table}[h]
\caption{Ablation Experiments of Various Modules of CSR}
\label{table2}
\begin{center}
\setlength\tabcolsep{3.5pt}
\begin{tabular}{ccccccccc}
\toprule
\upshape{Dataset} &\multicolumn{3}{c}{CIFAR-10}&\makecell{10N}& \multicolumn{3}{c}{CIFAR-100} & \makecell{100N}\\
\midrule
 Noise type & \multicolumn{3}{c}{Instance dependent} & \makecell{Worst}&\multicolumn{3}{c}{Instance dependent} & \makecell{Noise \\ fine}\\
\midrule
Noise rate & 0.2 & 0.4 & 0.6 &\makecell{about \\ 0.4}& 0.2 & 0.4 & 0.6 & \makecell{about \\ 0.4}  \\
\midrule
CSR &\textbf{94.08}	&\textbf{92.56} &\textbf{74.69}& \textbf{87.37}&\textbf{75.93}	&\textbf{71.75}	&\textbf{58.98}	&\textbf{64.62}\\
CSR w/o cw& 94.04& 92.07& 73.24&86.78&75.57 &71.57	&56.15	&63.09\\
CSR w/o $M$ & 93.95& 92.27&75.21&87.12&74.38&68.62	&53.94&	63.24\\
SOP & 93.91&91.95& 72.12& 86.76&74.09&66.52	&50.20	&61.09\\
\bottomrule
\end{tabular}
\end{center}
\end{table}

\subsubsection{Ablation Study of CSR}

 CSR adds the collaboration matrix and confidence weight to SOP. In order to further evaluate the respective contributions of these two modules, we conducted an ablation study on CSR. Table \ref{table2} presents the results of this study. It becomes evident that the removal of either module significantly impairs performance. However, the elimination of the collaboration matrix has a more detrimental effect on performance compared to the removal of the confidence weights. This observation further aligns with our previous findings. The collaboration matrix effectively diminishes generalization errors by mitigating erroneous gradient leakage, while the confidence weight enables the network to learn and generate more reliable predictions. When used together, the test accuracy can be maximized. 

We believe that either the collaboration matrix or the confidence weight holds the potential to effectively mitigate overfitting to noisy labels in the model. To validate our hypothesis, we have introduced a novel evaluation metric termed Noise Fitting Rate (NFR). This metric provides a quantitative measure of the samples in which the predictions accurately align with incorrect labels, relative to the overall number of mislabeled samples. Assuming  $\Omega_e$ is the collection of mislabeled samples, its calculation is defined by the following formula:
\begin{equation}
\label{e22}
NFR=| \ \{i|f_i(x_i,\theta)=y_i,i\in \Omega_{e}\} \ |\ /|\Omega_{e}|.
\end{equation}

Within this context, the variable $i$ represents the $i$-th individual sample within the sample collection $\Omega_e$. Meanwhile, $f_i(x_i,\theta)$ denotes the predicted result generated by the model for the $i$-th sample, and $y_i$ signifies the noisy label associated with that particular sample. Moreover, the notation $|\Omega_e|$ indicates the size of elements within the set $\Omega_e$. 

\begin{figure}[h]
\centering
\includegraphics[width=2.6in]{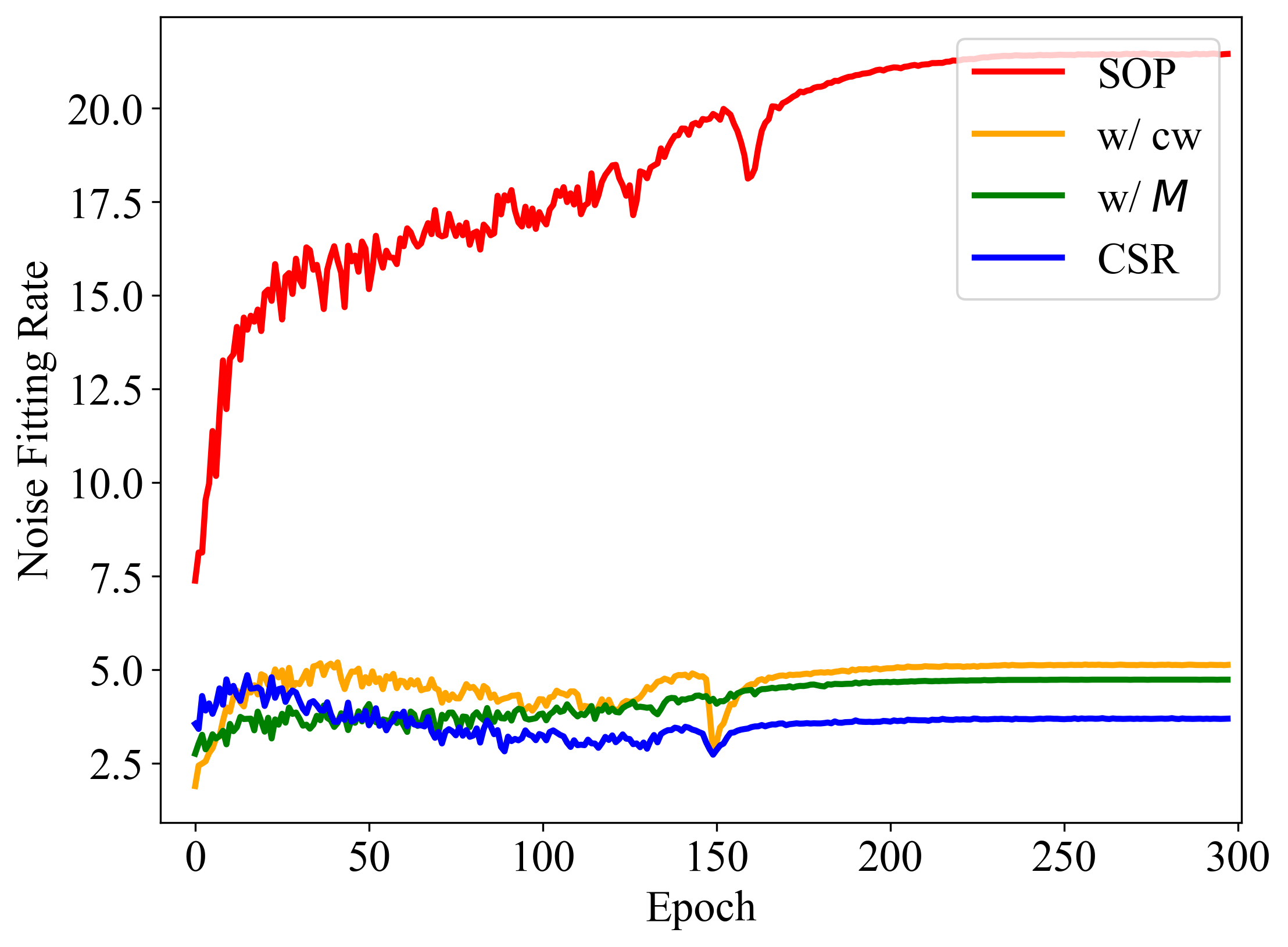}
\caption{Curves of Noise Fitting Rate over time on the CIFAR-100 40\% IDN dataset.}
\label{fig13}
\end{figure}

To visually illustrate the fluctuation of this Noise Fitting Rate (NFR) over time, we have constructed a graph, as depicted in Figure \ref{fig13}. This figure observes and demonstrates the level of noise fitting by methods in the ablation study under 40\% instance-dependent noise on CIFAR-100. It is apparent that the baseline method SOP, represented by the red curve, struggles to withstand the detrimental impact of label noise in the absence of the collaboration matrix and the confidence weight. As the training proceeds, the network tends to increasingly overfit the noisy data. Conversely, the green and yellow curves depict learning processes incorporating the collaboration matrix and the confidence weight, respectively. These approaches successfully mitigate the negative effects of label noise during the early stage of training and enable robust learning throughout the subsequent training process. The proposed CSR, represented by the blue curve, demonstrates the highest level of robustness against label noise.\\

\subsubsection{Choice of $\tau_m$ }

In order to make the collaboration matrix effective, we experimentally determine the best value of the learning rate $\tau_m$. We conducted experiments where we updated matrix $M$ using various learning rates under CIFAR-100 40\% IDN, observing the impact on test performance. We discovered that a learning rate of 0.001 yielded the best performance for $M$. We utilize similar experiments to determine the learning rate of the collaboration matrix on other datasets. Specifically, for CIFAR-10, we ultimately settled on a learning rate of 0.000001 for the collaboration matrix. On other real datasets, the learning rate $\tau_m$ is set to 0.001.

\begin{table}
\caption{Ablation Experiment of $\tau_m$ Under 40\% IDN on CIFAR-100}
\label{table3}
\begin{center}
\begin{tabular}{cccccc}
\toprule 
\upshape{lr for $M$} & 0.1 & 0.01 & 0.001 & 0.0001    & 0.00001 \\
\midrule
test acc & 42.35 & 69.88 & \textbf{71.75} & 69.33 & 68.79 \\
\bottomrule
\end{tabular}
\end{center}
\end{table}

\subsection{Comparison of CSR+ with SOTA}
We compare CSR+ with the state-of-the-art methods in the field of noisy label learning to evaluate the effectiveness of the proposed robust learning framework. To guarantee fairness, the comparative methods were tested using the most advanced configuration, referred to as the Plus version. The results of these comparisons are presented in Table \ref{table4} in chronological order, with certain outcomes derived from publicly accessible data mentioned in the literature \cite{li2023disc}.

\begin{table}[h]
\caption{Comparison of CSR+ With SOTA on the CIFAR-10/100 and CIFAR-10N/100N Datasets}
\label{table4}
\begin{center}
\setlength\tabcolsep{3pt}
\begin{tabular}{ccccccccc}
\toprule
\upshape{Dataset} & \multicolumn{3}{c}{CIFAR-10}& 10N& \multicolumn{3}{c}{CIFAR-100} & \makecell{100N}\\
\midrule
  Noise type & \multicolumn{3}{c}{Instance dependent} & Worst& \multicolumn{3}{c}{Instance dependent} & \makecell{Noise \\ fine}\\
\midrule
Noise rate & 0.2 & 0.4 & 0.6& \makecell{about \\ 0.4} & 0.2 & 0.4 & 0.6 & \makecell{about \\ 0.4}  \\
\midrule
Co-teaching+ & \makecell{89.80\\±0.28} & \makecell{73.38\\±1.39}& \makecell{59.22\\±6.34}& \makecell{83.83\\±1.28} &\makecell{41.71\\±0.78}& \makecell{24.45\\ ±0.71}	&\makecell{12.58\\±0.58}	&\makecell{60.37\\ ±0.27}\\
ELR+& \makecell{95.09\\±0.21}& \makecell{94.06\\±0.32}&	\makecell{91.22\\±0.51}& \makecell{90.91\\±0.28}& \makecell{77.09\\ ±0.20}& \makecell{74.50\\±0.32 }& \makecell{66.99\\±0.79 }&	\makecell{67.73\\±0.41 }\\
DivideMix& \makecell{93.24\\±0.12}& \makecell{94.37\\±0.22} &\makecell{92.19\\±0.63}& \makecell{92.56\\±0.42} & \makecell{79.04\\ ±0.21}&\makecell{76.08\\±0.35 }&\makecell{46.72\\ ±1.32}&\makecell{71.13\\±0.48 }	\\
UNICON&	\makecell{95.92\\±0.01} &\makecell{95.72\\±0.02} &\makecell{95.34\\±0.04} & \makecell{94.48\\±0.01} &\makecell{78.22\\±0.09 }& \makecell{77.71\\ ±0.10}&	\makecell{71.07\\ ±0.32}&\makecell{70.84\\ ±0.11}\\
SOP+ &\makecell{96.19\\±0.12} &\makecell{95.51\\±0.16}& \makecell{84.86\\±0.19} & \makecell{93.24\\±0.21} &\makecell{78.26\\ ±0.12}&	\makecell{72.79\\±0.24 }& \makecell{62.25\\ ±0.31}& \makecell{67.81\\±0.23 }\\
CC &\makecell{93.68\\±0.12} &\makecell{94.97\\±0.09} &\makecell{94.95\\±0.11} &- &\makecell{79.61\\±0.19}&\makecell{76.58\\±0.25}&\makecell{59.40\\±0.46}&-\\
ProMix&	\makecell{97.12\\±0.04}	&\makecell{96.56\\±0.12} & \makecell{94.15\\±0.23} &\makecell{95.61\\±0.15} &\makecell{81.84\\ ±0.14}	&\makecell{79.50\\±0.23}	&\makecell{72.88\\±0.41 }	&\makecell{72.68\\±0.21}\\
DISC &\makecell{96.48\\±0.04} &\makecell{95.94\\±0.04}  &\makecell{\textbf{95.05}\\±0.05} & \makecell{90.37\\±0.13}&\makecell{80.12\\±0.13} &\makecell{78.44\\±0.19 } &\makecell{69.57\\±0.14 }&	\makecell{69.51\\±0.21}\\
CSR+& \makecell{\textbf{97.25}\\±0.04}& \makecell{\textbf{96.88}\\±0.07}& \makecell{94.57\\±0.16}& \makecell{\textbf{95.66}\\±0.14} & \makecell{\textbf{82.44}\\±0.09} &\makecell{\textbf{81.10}\\±0.13}& \makecell{\textbf{75.85}\\±0.23}& \makecell{\textbf{72.74}\\±0.14}\\
\bottomrule
\end{tabular}
\end{center}
\end{table}

The mean and standard deviation of three independently run experiments are presented. It can be seen that CSR+ outperforms most of the compared state-of-the-art methods on CIFAR-10/100 with various ratios of noise as well as CIFAR-10N/100N, demonstrating good generalization. CSR+ not only performs exceptionally well on instances with low noise ratios, but it also maintains relatively high test accuracy on datasets with 60\% instance-dependent noise. In comparison, Co-teaching+ and SOP+ show significant performance decline as the noise ratio increases. The DISC method exhibits unstable performance, although it achieves the highest test accuracy on the CIFAR-10 60\% IDN dataset, even surpassing our CSR+ method by 0.48 percentage points. However, it experiences noticeable declines on the CIFAR-100 60\% IDN dataset and two real-world noise datasets. Its performance on real datasets is even inferior to DivideMix, proposed in 2020. The CC method does not provide results on real-world datasets, but it also performs poorly on CIFAR-100, indicating that the assumptions of such methods may not hold in more complex scenarios. Overall, CSR+ has the best comprehensive performance. It is effective not only on data with more categories, but also on data with higher ratios of IDN noise and real-world noise.

Table \ref{table5} presents the experimental results of CSR+ on real-world datasets such as Clothing1M, WebVision, and ILSVRC12. Similarly, in terms of classification accuracy, our method outperforms all other state-of-the-art methods in terms of top-1 test accuracy on all three datasets. This further demonstrates the wide applicability and effectiveness of CSR+ in handling label noise in real-world scenarios. However, in terms of top-5 accuracy, CSR+ slightly lags behind CC's method. This could be attributed to CC's training approach that maximizes representation consistency and minimizes classification consistency, whereas CSR+ only utilizes a consistency regularization term.
\begin{table}[h]
\caption{Comparison of CSR+ With SOTA on the Clothing1M, Webvision and ILSVRC12 Datasets}
\label{table5}
\begin{center}
\begin{tabular}{cccccc}
\toprule 
\multirow{2}{*}{\upshape{Method}} & \upshape{Clothing1M}&\multicolumn{2}{c}{\upshape{WebVision}} & \multicolumn{2}{c}{\upshape{ILSVRC12}}\\
 && \upshape{top1} &\upshape{top5} &\upshape{top1} &\upshape{top5}  \\
\midrule
\upshape{Co-teaching} &69.21 &63.58&85.20 &61.48 &84.70\\
ELR+ &74.81& 77.78 & 91.68 & 70.29 & 89.76 \\
\upshape{DivideMix} &74.76&77.32 & 91.64&75.20& 90.84 \\
CC &74.54 & 79.36 & \textbf{93.64} & 76.08 & \textbf{93.86} \\
DISC & 74.79& 80.28 & 92.28 & 77.44 & 92.28 \\
CSR+ & \textbf{74.92}& \textbf{80.47} & 92.51 & \textbf{77.46} & 95.28 \\
\bottomrule
\end{tabular}
\end{center}
\end{table}

\subsubsection{Ablation Study on Sample Selection Strategies}

Due to the use of various latest training techniques in comparative methods, it is hard to evaluate the effectiveness of our sample selection strategy based purely on the final leaderboard. In order to further assess the proposed joint sample selection strategy, we conducted ablation experiments on different sample selection strategies within the unified CSR+ framework. The representative sample selection strategies we employed included BARE's selection based on Batch statistics, GMM+loss selection in DivideMix, mutual information selection in Co-teaching, JS divergence selection in UNICON.

The outcomes of the experiments are presented in Table \ref{table6}.
The utilization of GMM+loss proves to be extremely efficient when the noise ratio is low. However, as the noise ratio increases, this selection strategy runs the risk of misclassifying erroneous samples as correct, therefore amplifying the potential for confirmation bias. In scenarios characterized by a multitude of categories, the accurate computation of class centers becomes challenging, thus impeding the preciseness of sample selection in methods relying on latent space distance metrics. Conversely, mutual information-based methods address this issue by training two independent networks, thereby mitigating the accumulation of errors to a certain extent. In comparison, the joint sample selection strategy presented in this paper is more effective and yields superior outcomes in comparison to methods based on mutual information selection.

\begin{table}[h]
\caption{Comparison of Joint Sample Selection With the Other Sample Selection Methods}
\label{table6}
\begin{center}
\begin{tabular}{ccccc}
\toprule
\upshape{Dataset} & \multicolumn{3}{c}{CIFAR-100} & \makecell{CIFAR-100N}\\
\midrule
  Noise type  & \multicolumn{3}{c}{Instance dependent} & \makecell{Noise fine}\\
\midrule
Noise rate & 0.2 & 0.4 & 0.6 & \makecell{about 0.4}  \\
\midrule
batch statistics &	73.78	&72.06&	67.95&	63.37\\
GMM+loss	&82.34	&78.63	&73.08&	71.25\\
mutual information &81.74	&79.56	&74.00&	71.14\\
JS divergence	&82.21	&78.83&	71.64	&70.41\\
joint sample selection& \textbf{82.44}	&\textbf{81.10}& \textbf{75.85}	&\textbf{72.74}\\

\bottomrule
\end{tabular}
\end{center}
\end{table}

\subsubsection{Ablation Study of CSR+}
\begin{table}[h]
\caption{Ablation Experiments on CSR+}
\label{table7}
\begin{center}
\begin{tabular}{ccccc}
\toprule 
\upshape{Dataset} & \multicolumn{3}{c}{CIFAR-100} & \makecell{CIFAR-100N}\\
\midrule
  Noise type  & \multicolumn{3}{c}{Instance dependent} & \makecell{Noise fine}\\
\midrule
Noise rate & 0.2 & 0.4 & 0.6 & \makecell{about 0.4}  \\
\midrule
CSR	&75.93 &71.75 &58.98&64.62\\
+ CR & 77.26&73.34&65.44&68.18\\
+ Mixup & 81.17&78.26&66.50&70.16\\
+ Label Correction	&81.94&79.60&71.45&71.45\\
+ Co-train	&\textbf{82.44}&\textbf{81.10}&\textbf{75.85}&\textbf{72.74} \\
\bottomrule
\end{tabular}
\end{center}
\end{table}

We conducted ablation experiments on the modules of CSR+ to observe the contributions of the training techniques employed. We used CSR as the baseline method and gradually introduced noise-robust techniques such as consistency regularization, Mixup, Label Correction, and Co-train. The experiments were performed on two types of noise, namely real noise on CIFAR-100N and instance noises on CIFAR-100. The results are presented in Table \ref{table7}. It can be observed that each noise-robust technique improved the testing accuracy of CSR, leading to the best overall performance. The joint sample selection strategy required the combination of consistency regularization and Mixup to be effective. From the ablation experiments, it is evident that these two modules provided the most significant improvement in testing accuracy. Label correction and Co-train techniques also increased the accuracy, albeit to a lesser extent. The ablation experiments suggest that the CSR+ framework is not merely a stack of training tricks, but a robust framework for learning under noisy labels built around CSR learning and joint sample selection.

\subsection{Complexity Analysis}
During our experiments on CIFAR-100 with 40\% IDN labels, we conducted complexity analysis, which included time complexity and space complexity. We tabulated the changes in time consumption and memory usage for each module added after the SOP. The second column represents the time taken to train one epoch, the third column denotes the total memory consumption during training, the fourth column indicates the number of parameters, and the fifth column displays the testing accuracy. It can be observed that, compared to SOP, CSR does not significantly increase training time and memory consumption. However, it led to a significant improvement in performance, with the testing accuracy increasing from 66.52 to 71.75, a gain of 5.23 percentage points. CSR+ utilizes more parameters and consumes more time and memory during training. Nonetheless, it can further enhance the performance by almost 10 percentage points.

\begin{table}[h]
\caption{Complexity Analysis}
\setlength\tabcolsep{3pt}
\label{table8}
\begin{center}
\begin{tabular}{ccccc}
\toprule 
\upshape{Method} & Time(s) & Memory(MB) & Parameter(MB) & Accuracy(\%)    \\
\midrule
SOP & 15.3 & 835.9 & 18.1000& 66.52\\
CSR w/ cw & 15.4 & 836.0 & 18.1009& 71.57\\
CSR w/ M & 15.8 & 856.0 & 18.1000& 68.62\\ 
CSR & 16.2 & 856.0 & 18.1009& 71.75\\
\hspace{0.07cm} CSR+ & 105.2 & 2433.4 & 36.2018& 81.10\\
\bottomrule
\end{tabular}
\end{center}
\end{table}

\section{Conclusion}
Label noise is widely present in datasets and can significantly harm the generalization performance of models. Particularly in over-parameterized networks, it can also lead to a greater demand for more data, longer training duration, and larger network parameters. To address this issue, this paper proposes a coordinated sparse recovery (CSR) method based on sparse modeling of noise and over-parameterized training. This method resolves the inherent and widespread learning incoordination between model prediction and noise recovery. Furthermore, we extend this method to build a complete and robust learning framework. Extensive comparative and ablation experiments are conducted, which demonstrate that the proposed CSR method indeed reduces erroneous gradient updates, makes better use of early learning, and achieves improved generalization performance. The joint sample selection we construct is a highly effective strategy for sample selection, which, when combined with common semi-supervised training techniques, can achieve the best test accuracy on datasets with a higher proportion of instance-level noise. In the future, our research will focus on expanding the proposed method to address the openset problem and exploring better integration with pre-trained large-scale models.



\ifCLASSOPTIONcompsoc
  \section*{Acknowledgments}
\else
  \section*{Acknowledgment}
\fi

I would like to express my gratitude to Mr.JAHANZAIB YAQOOB for proofreading this manuscript, as well as to the laboratory staff for their valuable discussions and support in this endeavor.

\bibliographystyle{IEEEtran} 
\bibliography{IEEEabrv,main} 

\begin{thebibliography}{10}
\providecommand{\url}[1]{#1}
\csname url@samestyle\endcsname
\providecommand{\newblock}{\relax}
\providecommand{\bibinfo}[2]{#2}
\providecommand{\BIBentrySTDinterwordspacing}{\spaceskip=0pt\relax}
\providecommand{\BIBentryALTinterwordstretchfactor}{4}
\providecommand{\BIBentryALTinterwordspacing}{\spaceskip=\fontdimen2\font plus
\BIBentryALTinterwordstretchfactor\fontdimen3\font minus
  \fontdimen4\font\relax}
\providecommand{\BIBforeignlanguage}[2]{{%
\expandafter\ifx\csname l@#1\endcsname\relax
\typeout{** WARNING: IEEEtran.bst: No hyphenation pattern has been}%
\typeout{** loaded for the language `#1'. Using the pattern for}%
\typeout{** the default language instead.}%
\else
\language=\csname l@#1\endcsname
\fi
#2}}
\providecommand{\BIBdecl}{\relax}
\BIBdecl

\bibitem{zhang2021understanding}
C.~Zhang, S.~Bengio, M.~Hardt, B.~Recht, and O.~Vinyals, ``Understanding deep
  learning (still) requires rethinking generalization,'' \emph{Communications
  of the ACM}, vol.~64, no.~3, pp. 107--115, 2021.

\bibitem{arpit2017closer}
D.~Arpit, S.~Jastrz{\k{e}}bski, N.~Ballas, D.~Krueger, E.~Bengio, M.~S. Kanwal,
  T.~Maharaj, A.~Fischer, A.~Courville, Y.~Bengio \emph{et~al.}, ``A closer
  look at memorization in deep networks,'' in \emph{International conference on
  machine learning}.\hskip 1em plus 0.5em minus 0.4em\relax PMLR, 2017, pp.
  233--242.

\bibitem{jiang2018mentornet}
L.~Jiang, Z.~Zhou, T.~Leung, L.-J. Li, and L.~Fei-Fei, ``Mentornet: Learning
  data-driven curriculum for very deep neural networks on corrupted labels,''
  in \emph{International conference on machine learning}.\hskip 1em plus 0.5em
  minus 0.4em\relax PMLR, 2018, pp. 2304--2313.

\bibitem{song2022learning}
H.~Song, M.~Kim, D.~Park, Y.~Shin, and J.-G. Lee, ``Learning from noisy labels
  with deep neural networks: A survey,'' \emph{IEEE Transactions on Neural
  Networks and Learning Systems}, 2022.

\bibitem{natarajan2013learning}
N.~Natarajan, I.~S. Dhillon, P.~K. Ravikumar, and A.~Tewari, ``Learning with
  noisy labels,'' \emph{Advances in neural information processing systems},
  vol.~26, 2013.

\bibitem{manwani2013noise}
N.~Manwani and P.~Sastry, ``Noise tolerance under risk minimization,''
  \emph{IEEE transactions on cybernetics}, vol.~43, no.~3, pp. 1146--1151,
  2013.

\bibitem{van2015learning}
B.~Van~Rooyen, A.~Menon, and R.~C. Williamson, ``Learning with symmetric label
  noise: The importance of being unhinged,'' \emph{Advances in neural
  information processing systems}, vol.~28, 2015.

\bibitem{ghosh2017robust}
A.~Ghosh, H.~Kumar, and P.~S. Sastry, ``Robust loss functions under label noise
  for deep neural networks,'' in \emph{Proceedings of the AAAI conference on
  artificial intelligence}, vol.~31, no.~1, 2017.

\bibitem{zhang2018generalized}
Z.~Zhang and M.~Sabuncu, ``Generalized cross entropy loss for training deep
  neural networks with noisy labels,'' \emph{Advances in neural information
  processing systems}, vol.~31, 2018.

\bibitem{wang2019symmetric}
Y.~Wang, X.~Ma, Z.~Chen, Y.~Luo, J.~Yi, and J.~Bailey, ``Symmetric cross
  entropy for robust learning with noisy labels,'' in \emph{Proceedings of the
  IEEE/CVF international conference on computer vision}, 2019, pp. 322--330.

\bibitem{ma2020normalized}
X.~Ma, H.~Huang, Y.~Wang, S.~Romano, S.~Erfani, and J.~Bailey, ``Normalized
  loss functions for deep learning with noisy labels,'' in \emph{International
  conference on machine learning}.\hskip 1em plus 0.5em minus 0.4em\relax PMLR,
  2020, pp. 6543--6553.

\bibitem{wang2022scalable}
Y.~Wang, X.~Sun, and Y.~Fu, ``Scalable penalized regression for noise detection
  in learning with noisy labels,'' in \emph{Proceedings of the IEEE/CVF
  Conference on Computer Vision and Pattern Recognition}, 2022, pp. 346--355.

\bibitem{Bishop1995tikhonov}
C.~M. Bishop, ``Training with noise is equivalent to tikhonov regularization,''
  \emph{Neural Computation}, vol.~7, no.~1, pp. 108--116, 1995.

\bibitem{gu2008generative}
L.~Gu and T.~Kanade, ``A generative shape regularization model for robust face
  alignment,'' in \emph{Computer Vision--ECCV 2008: 10th European Conference on
  Computer Vision, Marseille, France, October 12-18, 2008, Proceedings, Part I
  10}.\hskip 1em plus 0.5em minus 0.4em\relax Springer, 2008, pp. 413--426.

\bibitem{hoffman2019robust}
J.~Hoffman, D.~A. Roberts, and S.~Yaida, ``Robust learning with jacobian
  regularization,'' \emph{arXiv preprint arXiv:1908.02729}, 2019.

\bibitem{xia2020part}
X.~Xia, T.~Liu, B.~Han, N.~Wang, M.~Gong, H.~Liu, G.~Niu, D.~Tao, and
  M.~Sugiyama, ``Part-dependent label noise: Towards instance-dependent label
  noise,'' \emph{Advances in Neural Information Processing Systems}, vol.~33,
  pp. 7597--7610, 2020.

\bibitem{yang2021estimating}
S.~Yang, E.~Yang, B.~Han, Y.~Liu, M.~Xu, G.~Niu, and T.~Liu, ``Estimating
  instance-dependent label-noise transition matrix using dnns,'' 2021.

\bibitem{arazo2019unsupervised}
E.~Arazo, D.~Ortego, P.~Albert, N.~O’Connor, and K.~McGuinness,
  ``Unsupervised label noise modeling and loss correction,'' in
  \emph{International conference on machine learning}.\hskip 1em plus 0.5em
  minus 0.4em\relax PMLR, 2019, pp. 312--321.

\bibitem{cheng2020learning}
H.~Cheng, Z.~Zhu, X.~Li, Y.~Gong, X.~Sun, and Y.~Liu, ``Learning with
  instance-dependent label noise: A sample sieve approach,'' \emph{arXiv
  preprint arXiv:2010.02347}, 2020.

\bibitem{englesson2021generalized}
E.~Englesson and H.~Azizpour, ``Generalized jensen-shannon divergence loss for
  learning with noisy labels,'' \emph{Advances in Neural Information Processing
  Systems}, vol.~34, pp. 30\,284--30\,297, 2021.

\bibitem{han2018co}
B.~Han, Q.~Yao, X.~Yu, G.~Niu, M.~Xu, W.~Hu, I.~Tsang, and M.~Sugiyama,
  ``Co-teaching: Robust training of deep neural networks with extremely noisy
  labels,'' \emph{Advances in neural information processing systems}, vol.~31,
  2018.

\bibitem{li2020dividemix}
J.~Li, R.~Socher, and S.~C. Hoi, ``Dividemix: Learning with noisy labels as
  semi-supervised learning,'' \emph{arXiv preprint arXiv:2002.07394}, 2020.

\bibitem{karim2022unicon}
N.~Karim, M.~N. Rizve, N.~Rahnavard, A.~Mian, and M.~Shah, ``Unicon: Combating
  label noise through uniform selection and contrastive learning,'' in
  \emph{Proceedings of the IEEE/CVF Conference on Computer Vision and Pattern
  Recognition}, 2022, pp. 9676--9686.

\bibitem{li2021learning}
J.~Li, C.~Xiong, and S.~C. Hoi, ``Learning from noisy data with robust
  representation learning,'' in \emph{Proceedings of the IEEE/CVF International
  Conference on Computer Vision}, 2021, pp. 9485--9494.

\bibitem{liu2020early}
S.~Liu, J.~Niles-Weed, N.~Razavian, and C.~Fernandez-Granda, ``Early-learning
  regularization prevents memorization of noisy labels,'' \emph{Advances in
  neural information processing systems}, vol.~33, pp. 20\,331--20\,342, 2020.

\bibitem{tanaka2018joint}
D.~Tanaka, D.~Ikami, T.~Yamasaki, and K.~Aizawa, ``Joint optimization framework
  for learning with noisy labels,'' in \emph{Proceedings of the IEEE conference
  on computer vision and pattern recognition}, 2018, pp. 5552--5560.

\bibitem{yu2019does}
X.~Yu, B.~Han, J.~Yao, G.~Niu, I.~Tsang, and M.~Sugiyama, ``How does
  disagreement help generalization against label corruption?'' in
  \emph{International Conference on Machine Learning}.\hskip 1em plus 0.5em
  minus 0.4em\relax PMLR, 2019, pp. 7164--7173.

\bibitem{zhang2021learning}
Y.~Zhang, S.~Zheng, P.~Wu, M.~Goswami, and C.~Chen, ``Learning with
  feature-dependent label noise: A progressive approach,'' \emph{arXiv preprint
  arXiv:2103.07756}, 2021.

\bibitem{zhu2022robustness}
Z.~Zhu, F.~Liu, G.~Chrysos, and V.~Cevher, ``Robustness in deep learning: The
  good (width), the bad (depth), and the ugly (initialization),''
  \emph{Advances in Neural Information Processing Systems}, vol.~35, pp.
  36\,094--36\,107, 2022.

\bibitem{wei2021open}
H.~Wei, L.~Tao, R.~Xie, and B.~An, ``Open-set label noise can improve
  robustness against inherent label noise,'' \emph{Advances in Neural
  Information Processing Systems}, vol.~34, pp. 7978--7992, 2021.

\bibitem{wang2019imae}
X.~Wang, Y.~Hua, E.~Kodirov, and N.~M. Robertson, ``Imae for noise-robust
  learning: Mean absolute error does not treat examples equally and gradient
  magnitude's variance matters,'' \emph{arXiv preprint arXiv:1903.12141}, 2019.

\bibitem{xu2019l_dmi}
Y.~Xu, P.~Cao, Y.~Kong, and Y.~Wang, ``L\_dmi: A novel information-theoretic
  loss function for training deep nets robust to label noise,'' \emph{Advances
  in neural information processing systems}, vol.~32, 2019.

\bibitem{goldberger2016training}
J.~Goldberger and E.~Ben-Reuven, ``Training deep neural-networks using a noise
  adaptation layer,'' in \emph{International conference on learning
  representations}, 2016.

\bibitem{xia2019anchor}
X.~Xia, T.~Liu, N.~Wang, B.~Han, C.~Gong, G.~Niu, and M.~Sugiyama, ``Are anchor
  points really indispensable in label-noise learning?'' \emph{Advances in
  neural information processing systems}, vol.~32, 2019.

\bibitem{patrini2017making}
G.~Patrini, A.~Rozza, A.~Krishna~Menon, R.~Nock, and L.~Qu, ``Making deep
  neural networks robust to label noise: A loss correction approach,'' in
  \emph{Proceedings of the IEEE conference on computer vision and pattern
  recognition}, 2017, pp. 1944--1952.

\bibitem{tanno2019learning}
R.~Tanno, A.~Saeedi, S.~Sankaranarayanan, D.~C. Alexander, and N.~Silberman,
  ``Learning from noisy labels by regularized estimation of annotator
  confusion,'' in \emph{Proceedings of the IEEE/CVF conference on computer
  vision and pattern recognition}, 2019, pp. 11\,244--11\,253.

\bibitem{malach2017decoupling}
E.~Malach and S.~Shalev-Shwartz, ``Decoupling" when to update" from" how to
  update",'' \emph{Advances in neural information processing systems}, vol.~30,
  2017.

\bibitem{liu2021co}
J.~Liu, R.~Li, and C.~Sun, ``Co-correcting: noise-tolerant medical image
  classification via mutual label correction,'' \emph{IEEE Transactions on
  Medical Imaging}, vol.~40, no.~12, pp. 3580--3592, 2021.

\bibitem{berthelot2019mixmatch}
D.~Berthelot, N.~Carlini, I.~Goodfellow, N.~Papernot, A.~Oliver, and C.~A.
  Raffel, ``Mixmatch: A holistic approach to semi-supervised learning,''
  \emph{Advances in neural information processing systems}, vol.~32, 2019.

\bibitem{patel2023adaptive}
D.~Patel and P.~Sastry, ``Adaptive sample selection for robust learning under
  label noise,'' in \emph{Proceedings of the IEEE/CVF Winter Conference on
  Applications of Computer Vision}, 2023, pp. 3932--3942.

\bibitem{li2023disc}
Y.~Li, H.~Han, S.~Shan, and X.~Chen, ``Disc: Learning from noisy labels via
  dynamic instance-specific selection and correction,'' in \emph{Proceedings of
  the IEEE/CVF Conference on Computer Vision and Pattern Recognition}, 2023,
  pp. 24\,070--24\,079.

\bibitem{nakkiran2021deep}
P.~Nakkiran, G.~Kaplun, Y.~Bansal, T.~Yang, B.~Barak, and I.~Sutskever, ``Deep
  double descent: Where bigger models and more data hurt,'' \emph{Journal of
  Statistical Mechanics: Theory and Experiment}, vol. 2021, no.~12, p. 124003,
  2021.

\bibitem{wen2022benign}
K.~Wen, J.~Teng, and J.~Zhang, ``Benign overfitting in classification: Provably
  counter label noise with larger models,'' in \emph{The Eleventh International
  Conference on Learning Representations}, 2022.

\bibitem{nguyen2022memorization}
D.~A. Nguyen, R.~Levie, J.~Lienen, G.~Kutyniok, and E.~H{\"u}llermeier,
  ``Memorization-dilation: Modeling neural collapse under noise,'' \emph{arXiv
  preprint arXiv:2206.05530}, 2022.

\bibitem{liu2022sop}
S.~Liu, Z.~Zhu, Q.~Qu, and C.~You, ``Robust training under label noise by
  over-parameterization,'' in \emph{International Conference on Machine
  Learning}.\hskip 1em plus 0.5em minus 0.4em\relax PMLR, 2022, pp.
  14\,153--14\,172.

\bibitem{you2020robust}
C.~You, Z.~Zhu, Q.~Qu, and Y.~Ma, ``Robust recovery via implicit bias of
  discrepant learning rates for double over-parameterization,'' \emph{Advances
  in Neural Information Processing Systems}, vol.~33, pp. 17\,733--17\,744,
  2020.

\bibitem{wani2023combining}
F.~A. Wani, M.~S. Bucarelli, and F.~Silvestri, ``Combining distance to class
  centroids and outlier discounting for improved learning with noisy labels,''
  \emph{arXiv preprint arXiv:2303.09470}, 2023.

\bibitem{chi-square}
X.~C. F. Z. Y.~L. Wei~Yang, Luhui~Xu, ``Chi-squared distance metric learning
  for histogram data,'' \emph{Mathematical Problems in Engineering}, 2015.

\bibitem{wang2022promix}
H.~Wang, R.~Xiao, Y.~Dong, L.~Feng, and J.~Zhao, ``Promix: combating label
  noise via maximizing clean sample utility,'' \emph{arXiv preprint
  arXiv:2207.10276}, 2022.

\bibitem{krizhevsky2009cifar100}
A.~Krizhevsky, G.~Hinton \emph{et~al.}, ``Learning multiple layers of features
  from tiny images,'' 2009.

\bibitem{wei2021cifar100n}
J.~Wei, Z.~Zhu, H.~Cheng, T.~Liu, G.~Niu, and Y.~Liu, ``Learning with noisy
  labels revisited: A study using real-world human annotations,'' \emph{arXiv
  preprint arXiv:2110.12088}, 2021.

\bibitem{xiao2015clothing1m}
T.~Xiao, T.~Xia, Y.~Yang, C.~Huang, and X.~Wang, ``Learning from massive noisy
  labeled data for image classification,'' in \emph{Proceedings of the IEEE
  conference on computer vision and pattern recognition}, 2015, pp. 2691--2699.

\bibitem{li2017webvision}
W.~Li, L.~Wang, W.~Li, E.~Agustsson, and L.~Van~Gool, ``Webvision database:
  Visual learning and understanding from web data,'' \emph{arXiv preprint
  arXiv:1708.02862}, 2017.

\bibitem{de2005ce}
P.-T. De~Boer, D.~P. Kroese, S.~Mannor, and R.~Y. Rubinstein, ``A tutorial on
  the cross-entropy method,'' \emph{Annals of operations research}, vol. 134,
  pp. 19--67, 2005.

\bibitem{willmott2005mae}
C.~J. Willmott and K.~Matsuura, ``Advantages of the mean absolute error (mae)
  over the root mean square error (rmse) in assessing average model
  performance,'' \emph{Climate research}, vol.~30, no.~1, pp. 79--82, 2005.

\end{thebibliography}
\vspace{-1.2cm}
\begin{IEEEbiography}[{\includegraphics[width=1in,height=1.25in,clip,keepaspectratio]{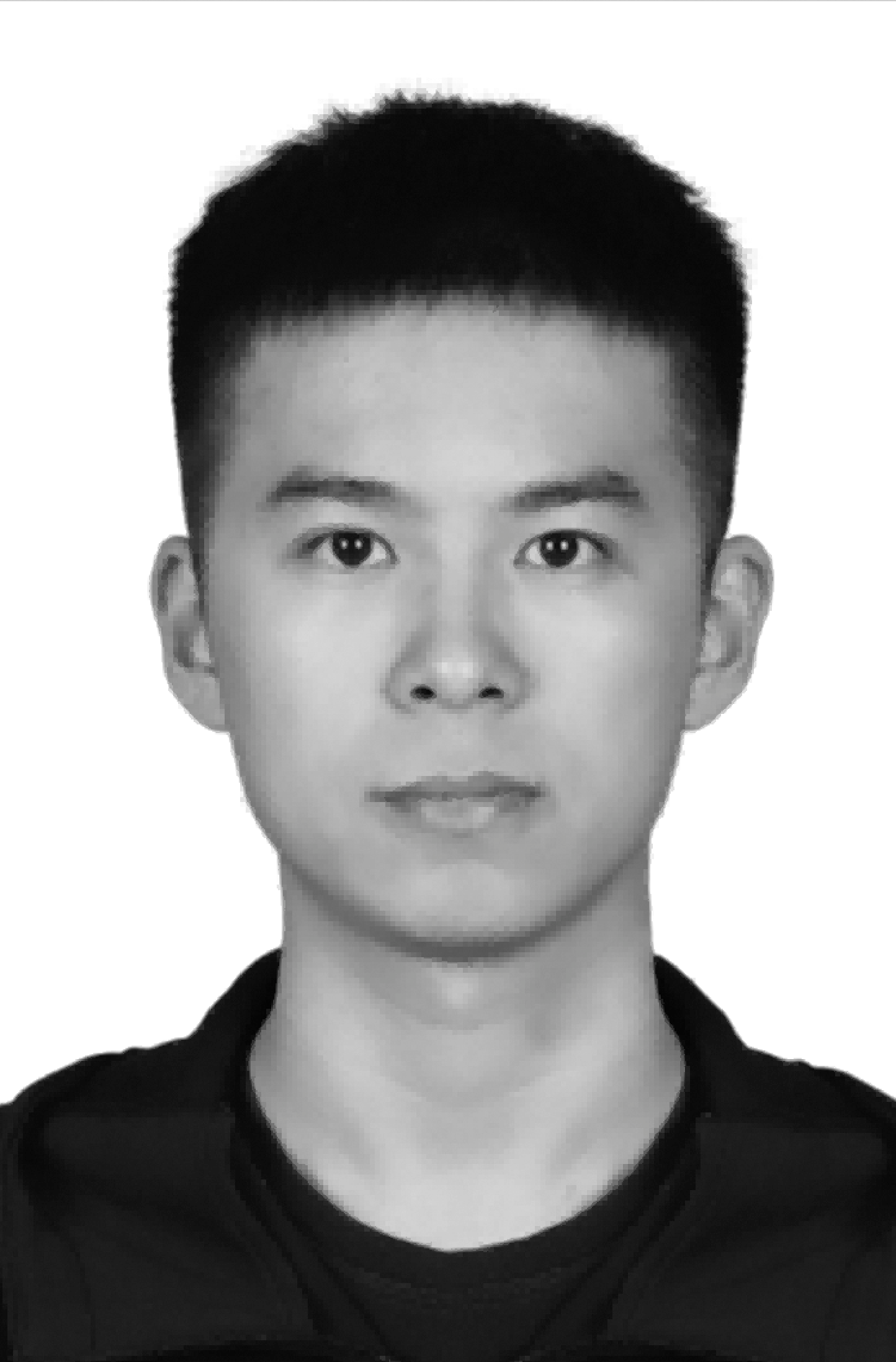}}]{Yukun Yang} received the BE degree from Qingdao University, China. He is currently working toward the MS degree with the Department of Information Science and Technology, Beijing University of Chemical Technology. His research interests include deep learning, robust learning and computer vision.
\end{IEEEbiography}
\vspace{-1.2cm}
\begin{IEEEbiography}[{\includegraphics[width=1in,height=1.25in,clip,keepaspectratio]{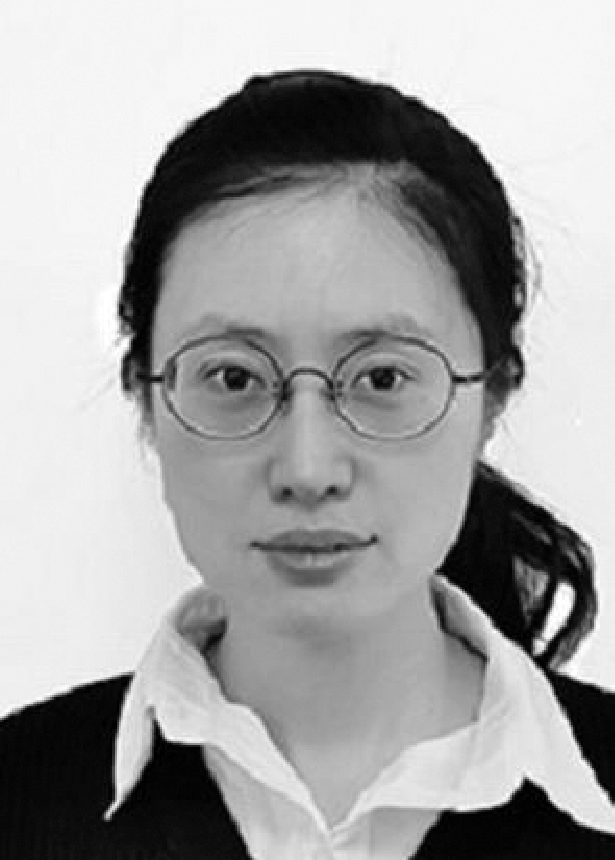}}]{Ruirui Li}
received the Ph.D. degree in computer science and technology from Tsinghua University, Beijing, China, in 2014. She joined as a Postdoctoral with the College of Information Science and Technology, Beijing University of Chemical Technology. She was appointed as an Associate Professor with the same institute in 2020. Her research interests include machine learning, pattern analysis, and computer vision.
\end{IEEEbiography}
\vspace{-1.2cm}
\begin{IEEEbiography}[{\includegraphics[width=1in,height=1.25in,clip,keepaspectratio]{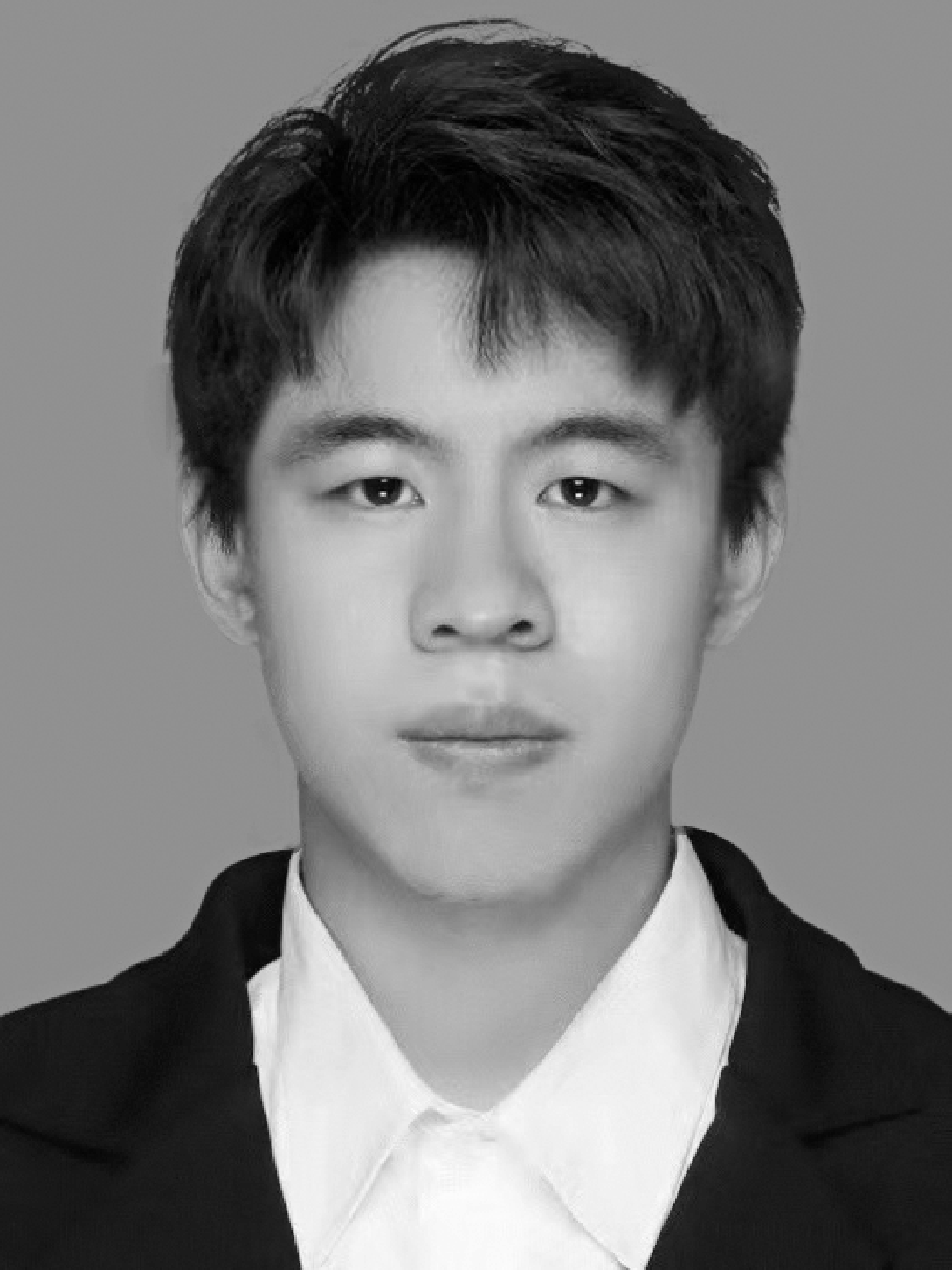}}]{Naihao Wang} received the BE degree from Beijing University of Chemical Technology, China. He is currently working toward the MS degree with the Department of Information Science and Technology, Beijing University of Chemical Technology. His research interests include deep learning, robust learning and computer vision.
\end{IEEEbiography}
\vspace{-1.2cm}
\begin{IEEEbiography}[{\includegraphics[width=1in,height=1.25in,clip,keepaspectratio]{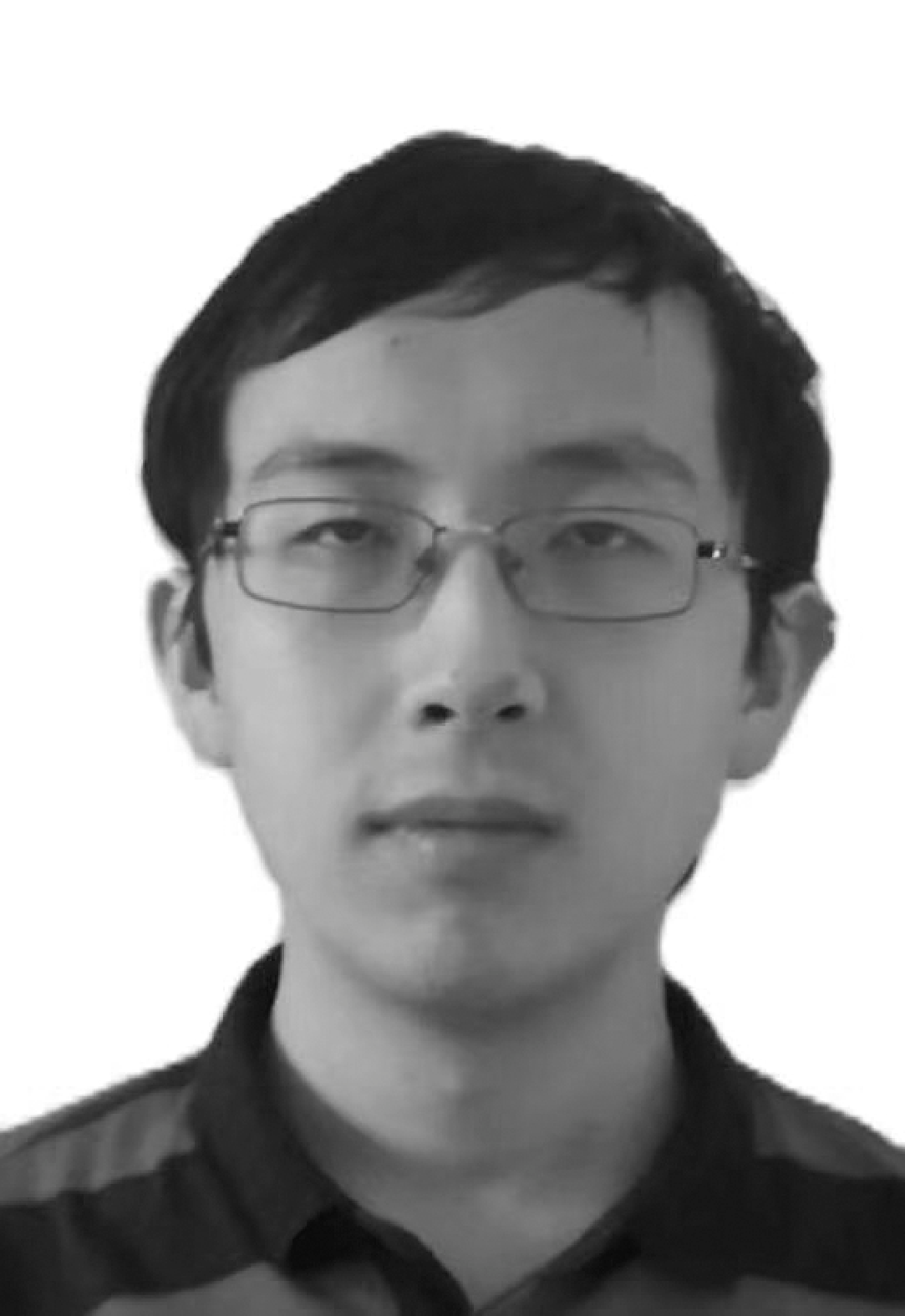}}]{Haixin Yang} received the BE degree from Peking University, China. He is currently working toward the Ph.D. degree with the School of Mathematical Sciences, Peking University. His research interests include deep learning, robust learning and computer vision.
\end{IEEEbiography}

\end{document}